\documentclass[10pt]{article} 
\usepackage[preprint]{tmlr}

\usepackage{amsmath,amsfonts,bm}

\def\eqref#1{equation~\ref{#1}}

\def\1{\bm{1}}

\DeclareMathAlphabet{\mathsfit}{\encodingdefault}{\sfdefault}{m}{sl}
\SetMathAlphabet{\mathsfit}{bold}{\encodingdefault}{\sfdefault}{bx}{n}
\newcommand{\tens}[1]{\bm{\mathsfit{#1}}}

\def\tE{{\tens{E}}}

\def\tX{{\tens{X}}}

\usepackage{hyperref}
\usepackage{url}

\usepackage{wrapfig}
\usepackage{graphicx}
\usepackage[noabbrev,capitalize]{cleveref}
\usepackage[ruled,vlined]{algorithm2e}
\usepackage{caption}
\usepackage{subcaption}
\usepackage{mathtools}
\usepackage[figuresleft]{rotating}
\usepackage{bbm}

\definecolor{mycolor}{RGB}{83,83,182}
\hypersetup{
    colorlinks=true,
    citecolor=mycolor,
    linkcolor=mycolor
}
\definecolor{new}{RGB}{0,0,0}  
\definecolor{camera}{RGB}{0,0,0} 
\definecolor{rebutal}{RGB}{0,0,0} 
\definecolor{tmlrrebutal}{RGB}{0,0,0} 

\newcommand{\redu}[1]{#1^\mathfrak{r}}
\newcommand{\rebutal}[1]{{\color{rebutal} #1}}
\newcommand{\tmlrrebutal}[1]{{\color{tmlrrebutal} #1}}

\title{PAVI:\\Plate-Amortized Variational Inference}

\author{%
    \name Louis Rouillard \email louis.rouillard-odera@inria.fr\\
    \addr Université Paris-Saclay, Inria, CEA\\
    Palaiseau, 91120, France
    \AND
    \name Alexandre Le Bris \email alexandre.le-bris@inria.fr\\
    \addr Université Paris-Saclay, Inria, CEA\\
    Palaiseau, 91120, France
    \AND
    \name Thomas Moreau \email thomas.moreau@inria.fr\\
    \addr Université Paris-Saclay, Inria, CEA\\
    Palaiseau, 91120, France
    \AND
    \name Demian Wassermann \email demian.wassermann@inria.fr\\
    \addr Université Paris-Saclay, Inria, CEA\\
    Palaiseau, 91120, France
}

\def\month{MM}  
\def\year{YYYY} 
\def\openreview{\url{https://openreview.net/forum?id=XXXX}} 

\begin{document}

\maketitle

\begin{abstract}
    Given observed data and a probabilistic generative model, Bayesian inference searches for the distribution of the model's parameters that could have yielded the data.
Inference is challenging for large population studies where millions of measurements are performed over a cohort of hundreds of subjects, resulting in a massive parameter space.
This large cardinality renders off-the-shelf Variational Inference (VI) computationally impractical.
In this work, we design structured VI families that efficiently tackle large population studies.
Our main idea is to share the parameterization and learning across the different i.i.d. variables in a generative model, symbolized by the model's \textit{plates}.
We name this concept \textit{plate amortization}.
Contrary to off-the-shelf stochastic VI, which slows down inference, plate amortization results in orders of magnitude faster to train variational distributions.
Applied to large-scale hierarchical problems, PAVI yields expressive, parsimoniously parameterized VI with an affordable training time.
This faster convergence effectively unlocks inference in those large regimes.
We illustrate the practical utility of PAVI through a challenging Neuroimaging example featuring 400 million latent parameters, demonstrating a significant step towards scalable and expressive Variational Inference.    
\end{abstract}

\section{Introduction}
Inference in graphical models allows to explain data in an interpretable and uncertainty-aware manner~\citep{koller_probabilistic_2009, Gelman_book, zhang_advances_2019}.
Such graphical models are leveraged in Neuroimaging to capture complex phenomena, for instance, the localization of cognitive function across the human brain cortex~\citep{parc_1, kong_spatial_2018}.
Yet, current inference methods struggle with the high dimensionality of Neuroimaging applications, which can feature hundreds of millions of latent parameters.
This work unlocks inference for those large-scale applications via novel scalable VI methodologies.

Our target applications are population studies.
Population studies analyze measurements over large cohorts of subjects.
They allow to model population parameters ---for instance, the mean weight in a population and different groups of subjects--- along with individual variability ---the variation of each subject's weight compared to the group weight.
These studies are ubiquitous in health care \citep{fayaz2016prevalence,towsley2011population}, and can typically involve hundreds of subjects and individual measurements per subject.
For instance, in Neuroimaging~\citep[e.g.][]{kong_spatial_2018}, measurements $X$ can correspond to signals in thousands of locations in the brain for a thousand subjects.
Given this observed data $X$, and a generative model that can produce $X$ given model parameters $\Theta$, we want to recover the {\color{new} parameters} $\Theta$ that could have yielded the observed $X$.
In our Neuroimaging example, $\Theta$ can be local labels for each brain location and subject, together with global parameters common to all subjects, such as the connectivity fingerprint corresponding to each label.
In many real-world applications, the link between latent parameters and observed data is not deterministic.
For instance, the observed signal can be noisy, or the modeled process in itself may be stochastic~\citep{koller_probabilistic_2009}.
As such, several parameters $\Theta$ instances could yield the same data $X$.
We want to model that parameter uncertainty by recovering the \textit{distribution} of the $\Theta$ that could have produced $X$.
Following the Bayesian formalism~\citep{Gelman_book}, we cast both $\Theta$ and $X$ as random variables (RVs), and our goal is to recover the \textit{posterior} distribution $p(\Theta|X)$.
Due to the nested structure of our applications, we focus on the case where $p$ corresponds to a hierarchical Bayesian model (HBM) \citep{Gelman_book}.
In population studies, the multitude of subjects and measurements per subject implies a large dimensionality for both $\Theta$ and $X$.
This large dimensionality in turn creates computational hurdles that we tackle through our method.

Several inference methods for HBMs exist in the literature.
Earliest works resorted to Markov chain Monte Carlo \citep{koller_probabilistic_2009}, which become slower as dimensionality increases \citep{blei_variational_2017}.
Recent approaches coined Variational Inference (VI) cast the inference as an optimization problem~\citep{blei_variational_2017, zhang_advances_2019, VCD}.
In VI, inference reduces to choosing a variational family $\mathcal{Q}$ and finding inside that family the distribution $q(\Theta; \phi) \in \mathcal{Q}$ closest to the unknown posterior $p(\Theta | X)$.
Historically, VI required to manually derive and optimize over $\mathcal{Q}$, which remains an effective method where applicable~\citep{manual_VI}.
This \textit{manual} VI requires technical mastery.
The experimenter has to choose an appropriate family $\mathcal{Q}$ to approximate the unknown posterior closely.
At the same time, the experimenter must rely on properties such as conjugacy to derive an optimizing routine~\citep{manual_VI, ADVI}.
This means that time and effort must be spent not only on the \textit{modeling} part ---laying out hypothesis in a generative model--- but also on the \textit{inference} part ---inferring parameters given the generative model.
For many experimenters, manual VI thus features strong barriers to entry, both in time and technical skill~\citep{ADVI}. 
In contrast, we follow the idea of \textit{automatic} VI: deriving an efficient family $\mathcal{Q}$ directly from the HBM $p$.
Automatic VI reduces inference to the \textit{modeling} part only, simplifying the experimenter's research cycle~\citep{ADVI, ASVI, CF}.
We propose an automatic VI method computationally applicable to large population studies.

Our method relies on the pervasive idea of amortization in VI.
Amortization can be broadly defined as finding posteriors usable across multiple data points~\citep{zhang_advances_2019}.
As such, amortization can be interpreted as the meta-learning of the inference problem~\citep{ABML,meta_learning_amo_VI, meta_learning_hierch_struct}.
A particular example of meta-learning is Neural Processes, which share with our method the conditioning of a density estimator by the output of a permutation-invariant encoder~\citep{NPF_1, NPF_2, deep_sets}.
Though close in spirit to our work, meta-learning studies problems with a single hierarchy.
One-hierarchy cases correspond to models distinguishing local parameters, representing the subjects, from global parameters, representing the population~\citep{ABML,HIM}.
As a notable example, \citet{agrawal_amortized_2021} provide theoretical guarantees in the single-hierarchy case.
In contrast, our focus is rather computational.
We furthermore study generic HBMs with an arbitrary number of hierarchies ---population, group and subject--- to tackle large population studies efficiently.

Modern VI is effective in low-dimensional settings but does not scale up to large population studies, which can involve millions of random variables~\citep[e.g.][]{kong_spatial_2018}.
In this work, we identify and tackle two challenges to enable this scale-up.
A first challenge with scalability is a detrimental trade-off between expressivity and high-dimensionality~\citep{ADAVI}.
To reduce the inference gap, VI requires the variational family $\mathcal{Q}$ to contain distributions closely approximating $p(\Theta | X)$~\citep{blei_variational_2017}.
Yet the form of $p(\Theta | X)$ is usually unknown to the experimenter.
Instead of a lengthy search for a valid family, one can resort to universal density approximators: normalizing flows~\citep{papamakarios_normalizing_2019}.
But the cost for this generality is a heavy parameterization, and normalizing flows scale poorly with the dimensionality of $\Theta$.
As a result, in large population studies, the parameterization of normalizing flows becomes prohibitive.
To tackle this challenge, \citet{ADAVI} recently proposed, via the ADAVI architecture, to partially share the parameterization of normalizing flows across the hierarchies of a generative model.
We build upon this shared parameterization idea while improving the ADAVI architecture on several aspects: removing the mean-field approximation; extending from pyramidal to arbitrary graphs; generalizing to non-sample-amortized and stochastically trained schemes.
Critically, while ADAVI tackled the over-parameterization of VI in population studies, it still could not perform inference in large data regimes.
This is due to a second challenge with scalability: as the size of $\Theta$ increases, evaluating a single gradient over the entirety of an architecture's weights quickly requires too much memory and computation.
This second challenge can be overcome using stochastic VI~\citep[SVI,][]{SVI}, which subsamples the parameters $\Theta$ inferred for at each optimization step.
However, using SVI, the weights for the posterior for a local parameter $\theta \in \Theta$ are only updated when the algorithm visits $\theta$.
In the presence of hundreds of thousands of such local parameters, stochastic VI can become prohibitively slow.

This work introduces the concept of \textit{plate amortization} (PAVI) for fast inference in large-scale HBMs.
Instead of considering the inference over local parameters $\theta$ as separate problems, our main idea is to share both the parameterization and learning across those local parameters, or equivalently across a model's \textit{plates}.
We first propose an algorithm to automatically derive an expressive yet parsimoniously-parameterized variational family from a plate-enriched HBM.
We then propose a hierarchical stochastic optimization scheme to train this architecture efficiently.
PAVI leverages the repeated structure of plate-enriched HBMs via a novel combination of amortization and stochastic training.
Through this combination, our main claim is to enable inference over arbitrarily large population studies with reduced parameterization and training time as the cardinality of the problem augments.
Critically, while traditional Stochastic VI unlocked large-scale inference at the cost of slower convergence, PAVI does not feature such a detrimental trade-off.
On the contrary, PAVI converges orders of magnitude faster than non-stochastic VI in large regimes.
In practice, this quicker training unlocks inference in applications previously unattainable, bringing inference time down from weeks to hours.
We illustrate this by applying PAVI to a challenging human brain cortex parcellation, featuring inference over a cohort of $1000$ subjects with tens of thousands of measurements per subject, for a total of half a billion RVs.
This demonstrates a significant step towards scalable, expressive and fast VI.
\section{Problem statement: inference in large population studies}
\label{sec:problem_statement}

Here we introduce a motivating example for our work that we'll use as an illustration in the rest of this section.
The concepts introduced in this section will be more formally defined below in \cref{sec:HBMs}.

\begin{figure}
    \centering
    \includegraphics[width=\textwidth]{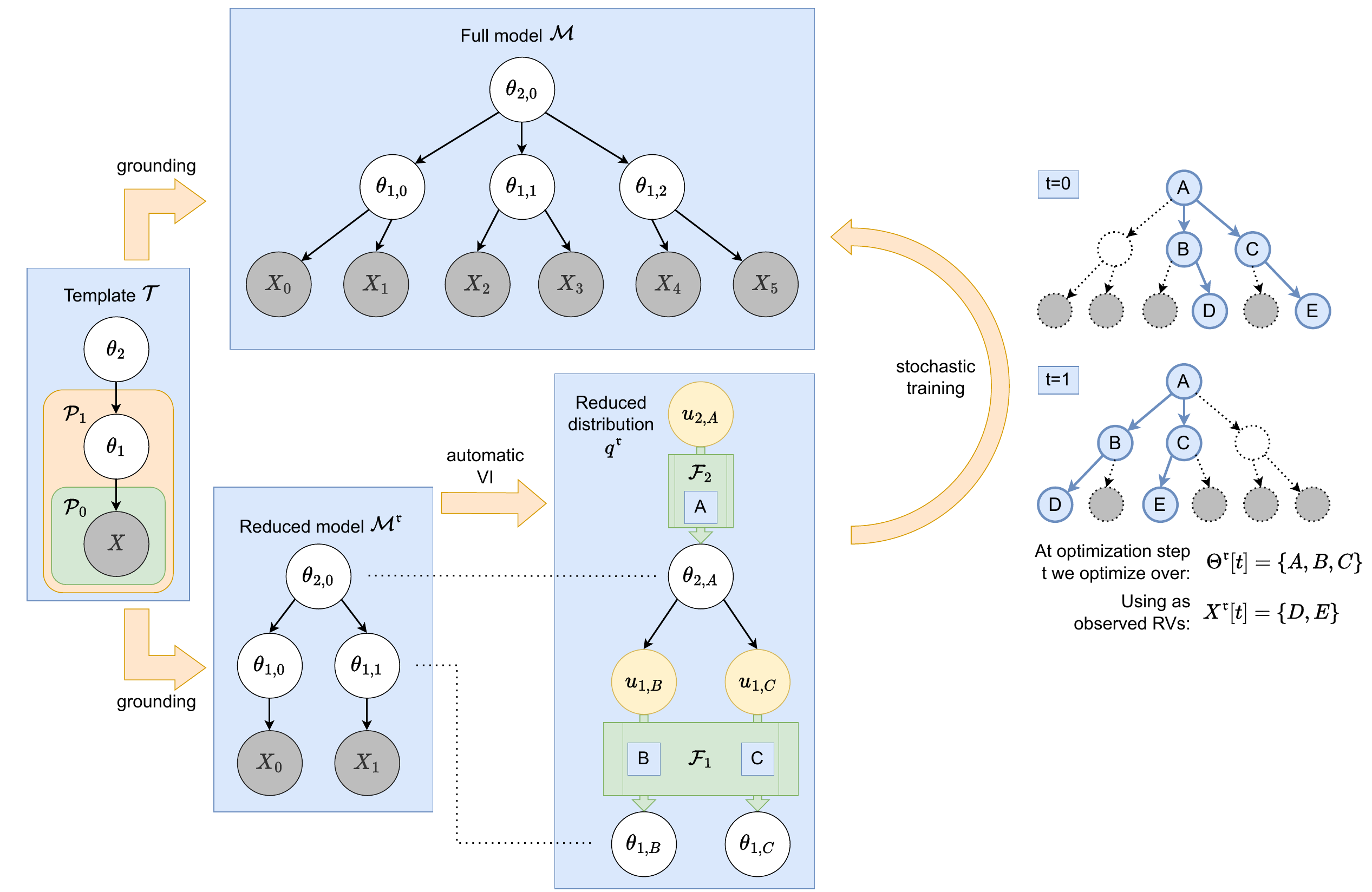}
    \caption{
        \tmlrrebutal{\textbf{PAVI working principle:}
        The template $\mathcal{T}$ (left) can be \textit{grounded} into the \textit{full} model $\mathcal{M}$ (top).
        We aim to perform inference over $\mathcal{M}$.
        Yet, $\mathcal{M}$ can feature large cardinalities in population studies.
        As an example, instead of $\theta_{1,0},..., \theta_{1,2}$, $\mathcal{M}$ can feature $\theta_{1,0}, ..., \theta_{1,1000}$ --corresponding to a thousand different subject groups.
        This can make inference over $\mathcal{M}$ computationally intractable.
        To circumvent this issue, we will train over $\mathcal{M}$ stochastically.
        To this end, we instantiate $\mathcal{M}$'s template into a smaller replica: the reduced model $\redu{\mathcal{M}}$ (bottom left).
        Following the automatic VI framework, we derive the reduced distribution $\redu{q}$ (bottom right) directly from $\redu{\mathcal{M}}$.
        The reduced distribution $\redu{q}$ features 2 conditional normalizing flows $\mathcal{F}_1$ and $\mathcal{F}_2$ respectively associated to the RV templates $\theta_1$ and $\theta_2$.
        During the stochastic training (right), $\redu{q}$ is instantiated over different branchings of the full model $\mathcal{M}$ --highlighted in blue.
        The branchings have $\redu{\mathcal{M}}$'s cardinalities and change at each stochastic training step $t$.
        The branching determine the encodings $\tE$ conditioning the flows $\mathcal{F}$ --as symbolised by the letters A, B, C-- and the observed data slice --as symbolised by the letters D, E.}
    }
    \label{fig:PAVI_principle}
    \label{fig:PAVI_training}
\end{figure}

Our objective is to perform inference in large population studies.
As an example of how inference becomes impractical in this context, consider $\mathcal{M}$ in \cref{fig:PAVI_principle} (top) as a model for the weight distribution in a population.
$\theta_{2,0}$ denotes the mean weight across the population.
$\theta_{1,0}, \theta_{1,1}, \theta_{1,2}$ denote the mean weights for $3$ groups of subjects, distributed around the population mean.
$X_0, X_1$ represent the observed weights of $2$ subjects from group $0$, distributed around the group mean.
Given the observed subject weights $X$, the goal is to determine the posterior distributions of the group and population means $p(\theta_{1, 0}, \theta_{1, 1}, \theta_{1, 2}, \theta_{2,0} | X)$.

To infer using the VI framework, we choose a variational family $\mathcal{Q}$ and search inside this family for the distribution closest to our unknown distribution of interest: $q(\Theta) \simeq p(\Theta | X)$.
Applying automatic VI to the example in \cref{fig:PAVI_principle}, the variational family $\mathcal{Q}$ will oftentimes factorize to~\citep{ASVI, CF}:
\begin{equation}
    \label{eq:example_facto}
    q(\Theta;\Phi) = q(\theta_{2,0} ;\phi_{2,0}) \prod_{n=0}^2 q(\theta_{1,n} | \theta_{2,0} ;\phi_{1,n})
\end{equation}
where $\Phi = \{ \phi_{2,0}, \phi_{1, 0}, ..., \phi_{1, 3} \}$ represent the weights associated to the variational family $\mathcal{Q}$ and each factor in $q$ will approximate the corresponding posterior distribution: as an example $q(\theta_{2,0}; \phi_{2,0}) \simeq p(\theta_{2, 0} | X)$.
Compared to the mean-field factorization~\citep{blei_variational_2017}, in \cref{eq:example_facto} the posterior for the RVs $\theta_{1,n}$ is conditional to the RV $\theta_{2,0}$ for greater expressivity.
We will follow this dependency scheme in \cref{sec:variational_family}.
During training, we will search for the optimal weights $\Phi$ to best approximate the posterior distribution $p(\Theta | X)$.

Yet, as the number of groups and subjects per group augments, this inference problem becomes computationally intractable.
On the parameterization side, each additional group in the population study requires additional weights $\phi_{1,n}$.
This constitutes a first computational hurdle: the total number of weights $\Phi$ in the variational family becomes prohibitively large.
On the computing side, optimizing over an increasing number of weights $\phi_{1,n}$ also requires a growing amount of calculation.
This constitutes a second computational hurdle.
To circumvent this issue, one must resort to Stochastic VI and only optimize over a subset of the weights $\Phi$ at a given training step.
In practice, this amounts to inferring the group weight of only a subset of the subject groups at a time.
Yet, stochastic training means that the weights $\phi_{1,n}$ corresponding to a given subject group will only be optimized for a fraction of the training time.
Consequently, the larger the population study becomes, the slower the inference.
Our goal is to keep inference computationally tractable as the cardinality of the problem augments.

To keep the inference tractable, we will build upon the natural factorization of the problem into subjects and groups of subjects.
This factorization is symbolized by a model's plates as detailed in \cref{sec:HBMs}.
First, we will harness this natural factorization in the parameterization of our variational family --to reduce its number of weights.
This strategy, coined \textit{plate amortization}, will be detailed in \cref{sec:plate_amortization}.
Second, we will reflect this factorization in the design of a stochastic training scheme --to control the computing required during the training.
This will be detailed in \cref{sec:training}.
Critically, our contribution does not lie only in adding those two items.
Combining shared parametrization and stochastic training, PAVI yields orders-of-magnitude speedups in inference.
Consequently, contrary to Stochastic VI, inference does not become prohibitively slow as the problem's cardinality augments.
In practice, this unlocks inference in very large population studies.
We illustrate this speed-up in our experiments in \cref{sec:exps} and the practical utility of our method in a challenging Neuroimaging setup in \cref{sec:exp_fMRI}.

\section{Generative modeling}
\label{sec:HBMs}
\subsection{hierarchical Bayesian models (HBMs), templates and plates}

\tmlrrebutal{
Here we define more formally the inference problem and the associated notations.

Population studies can be compactly represented via plate-enriched directed acyclic graph (DAG) templates $\mathcal{T}$ \citep[plates are defined in chapter 8.1 in][describes plates and template variables in chapter 6.3]{book_bishop, gilks_language_1994, koller_probabilistic_2009}.
$\mathcal{T}$ feature RV \textit{templates} that symbolise multiple similar \textit{ground} RVs.
As an example, the RV template $\theta_1$ in \cref{fig:PAVI_principle} (left) represents a generic group of subjects, of which there would be as many instances as the cardinality of the plate $\mathcal{P}_1$.
The plate structure in $\mathcal{T}$ also denotes that the multiple ground RVs corresponding to the same RV template are conditionally i.i.d.
As an example, two different group weights could be two Gaussian perturbations of the same population weight.
The two groups weight RVs thus have the same conditional distribution, and are independent given the population weight.
The plate structure in $\mathcal{T}$ thus symbolizes a strong symmetry in the problem that we exploit in our method.

We denote the template $\mathcal{T}$'s vertices, corresponding to RV templates, as $X$ and $\Theta=\{ \theta_i \}_{i=1 .. I}$.
$X$ denotes the RVs observed during inference, and $\Theta$ the parameters we infer: our goal is to approximate the posterior distribution $p(\Theta | X)$.
We denote $\mathcal{T}$'s plates as $\{ \mathcal{P}_p \}_{p=0 .. P}$, and the plates $\theta_i$ belongs to as $\operatorname{Plates}(\theta_i)$.
$I$ and $P$ respectively denote the number of latent RV templates and plates in $\mathcal{T}$, which are in general not equal.
In the toy example from \cref{fig:PAVI_principle}, there are two latent RV templates: $\theta_1$ and $\theta_2$, respectively the group and population mean weights.
$\mathcal{T}$ also features two plates $\mathcal{P}_1,\mathcal{P}_0$, which respectively denote groups in the population and the subjects in each group.
Graphically, we can see that $\operatorname{Plates}(\theta_2) = \emptyset$, whereas $\operatorname{Plates}(\theta_1) = \{ \mathcal{P}_1 \}$ and $\operatorname{Plates}(X) = \{ \mathcal{P}_0, \mathcal{P}_1 \}$.
To understand how we could have $P \neq I$, one could keep the plates $\mathcal{P}_1,\mathcal{P}_0$, but add two more RV templates to symbolize the population and group heights, in addition to their weights.
In this case, we would have $P=2$ and $I=4$.

By instantiating the repeated structures symbolized by the plates $\mathcal{P}$ in $\mathcal{T}$, we obtain a heavier graph representation: the hierarchical Bayesian model (HBM) $\mathcal{M}$.
This instantiation is visible in \cref{fig:PAVI_principle}, where we go from the template $\mathcal{T}$ (left) to the model $\mathcal{M}$ (top).
In the context of Neuroimaging~\citep{kong_spatial_2018}, $\mathcal{M}$ could feature millions of RVs, corresponding to thousands of subjects and measurements per subject, making it a much less compact representation.
We build upon the interplay between the template $\mathcal{T}$ and the HBM $\mathcal{M}$.
To go from one representation to the other, $\mathcal{T}$ can be \textit{grounded} into $\mathcal{M}$ given some plate cardinalities $\{ \operatorname{Card}(\mathcal{P}_p) \}_{p=0 .. P}$~\citep{koller_probabilistic_2009}.
$\operatorname{Card}(\mathcal{P})$ represents the number of elements in the plate $\mathcal{P}$, for instance the number of groups in the study.
Going from $\mathcal{T}$ to $\mathcal{M}$, a RV \textit{template} $\theta_i$ is instantiated into multiple \textit{ground} RVs $\{\theta_{i,n}\}_{n=0 .. N_i}$ with the same parametric form, where $N_i = \prod_{\mathcal{P} \in \operatorname{Plates}(\theta_i)} \operatorname{Card}(\mathcal{P})$.
Note that when RV templates belong to multiple plates, they symbolize as many ground RVs as the product of the cardinalities of those plates.
In \cref{fig:PAVI_principle}, the RV template $\theta_1$ is grounded into the ground RVs $\theta_{1, 0}, \theta_{1, 1}, \theta_{1, 2}$.
There are as many ground RVs $X_i$ as the product of the number of groups on the study $\operatorname{Card}(\mathcal{P}_1)$ times the number of subjects per group $\operatorname{Card}(\mathcal{P}_0)$.
We denote as $\pi(\theta_{i,n})$ the (potentially empty) set of parents of the RV $\theta_{i,n}$ in the ground graph corresponding to the model $\mathcal{M}$.
$\pi(\theta_{i,n})$ are the RVs whose value condition the distribution of $\theta_{i,n}$.
For instance, in \cref{fig:PAVI_principle}, a subject's weight ---the child RV--- is a perturbation of the group's weight ---the parent RV, conditioning the child RV's distribution.
This is denoted as $\pi(X_{0}) = \{ \theta_{1,0} \}$.

The benefit of the template $\mathcal{T}$ and the plates $\mathcal{P}$ is thus to symbolize the natural factorization of the graphical model $\mathcal{M}$ into i.i.d. RVs.
This factorization de-clutters the representation of hierarchical models.
PAVI exploits this simplifying factorization: we consider the inference over the ground RVs $\theta_{i,n}$ as symmetrical problems, over which we share parameterization and learning, as described in \cref{sec:methods}.
}

\subsection{Full model}

The \textit{full} model $\mathcal{M}$ is associated with the density $p$.
In $p$, the plate structure indicates that a RV template $\theta_i$ is associated to a conditional distribution $p_i$ shared across all ground RVs $\theta_{i,n}$:
 \begin{equation}
 \label{eq:p_full}
     \begin{aligned}
      \log p(\Theta, X) 
            &= log p(X | \Theta) + \log p(\Theta) \\
            &= \sum_{n=0}^{N_X} \log p_X(x_n | \pi(x_n)) + \sum_{i=1}^I \sum_{n=0}^{N_i} \log p_i(\theta_{i,n} | \pi(\theta_{i,n}))
     \end{aligned}
 \end{equation}
where $\pi(\theta_{i,n})$ is the (potentially empty) set of parents of the RV $\theta_{i,n}$, which condition its distribution.
We denote with a $\bullet_X$ index all variables related to the observed RVs $X$.
Exploiting the factorization visible in \cref{eq:p_full}, our goal is to obtain a variational distribution $q(\Theta)$ usable to approximate the unknown posterior $p(\Theta | X)$ for the target model $\mathcal{M}$.

\section{Methods}
\label{sec:methods}
\subsection{PAVI architecture}
\label{sec:archi}
 \paragraph{Full variational family design}
 
 \label{sec:variational_family}
 Here we define the variational distribution $q$ corresponding to the full model $\mathcal{M}$.
 To derive $q$, we push forward the prior $p(\Theta)$ using trainable normalizing flows, denoted as $\mathcal{F}$~\citep{papamakarios_normalizing_2019}.
 A normalizing flow is an invertible and differentiable neural network, also known as a diffeomorphism, that transforms samples from a base distribution into samples from a target distribution. 
To every ground RV $\theta_{i,n}$, we associate the learnable flow $\mathcal{F}_{i,n}$ to approximate its posterior distribution:
\begin{equation}
\label{eq:q_full}
    \begin{aligned}
    \log q(\Theta) &= \sum_{i=1}^I \sum_{n=0}^{N_i} \log q_{i,n}(\theta_{i,n} | \pi(\theta_{i,n})) \\
    \theta_{i,n} &= \mathcal{F}_{i,n}(u_{i,n}) \\
    u_{i, n} &\sim p_i(u_{i,n} | \pi(\theta_{i,n}))
    \end{aligned}
\end{equation}
where $q_{i,n}$ is the push-forward of the prior $p_i$ through the flow $\mathcal{F}_{i,n}$.
This push-forward is illustrated in \cref{fig:PAVI_training}, where flows $\mathcal{F}$ push RVs $u$ into $\theta$.
This \textit{cascading} scheme was first introduced by \citet{CF}, and makes $q$ inherit the conditional dependencies of the prior $p$.
As a result, $q(\Theta)$ in \cref{eq:q_full} mimics the right-hand term $p(\Theta)$ in \cref{eq:p_full}.
More details about the dependencies modeled in the variational distribution can be found in \cref{sec:dep_modelled}.


 \paragraph{Plate amortization}
 \label{sec:plate_amortization} 
 
 Here we introduce plate amortization: sharing the parameterization of density estimators across a model's plates.
 Plate amortization reduces the number of weights in a variational family as the cardinality of the inference problem augments.
In \cref{sec:sharing_learning}, we show that plate amortization also results in faster inference.
 
 VI searches for a distribution $q(\Theta ; \phi)$ that best approximates the posterior of $\Theta$ given a value $\tX_0$ for $X$: $q(\Theta ; \phi_0) \simeq p(\Theta | X=\tX_0)$.
 When presented with a new data point $\tX_1$, optimization has to be performed again to search for the weights $\phi_1$, such that $q(\Theta ; \phi_1) \simeq p(\Theta | X=\tX_1)$.
 Instead, \textit{sample amortized} (sa) inference \citep{zhang_advances_2019, amortization_gap} infers in the general case, regressing the weights $\phi$ using an \textit{encoder} $f$ of the observed data $\tX$: $q(\Theta ; \phi=f(\tX_i)) \simeq p(\Theta | X=\tX_i)$.
 The cost of learning the encoder weights is \textit{amortized} since inference over any new sample $\tX$ requires no additional optimization.
 In \cref{sec:exp_scaling}, we compare PAVI to several \textit{sample} amortized baselines, which will be denoted by the suffix \textit{(sa)}, and should not be confused with \textit{plate} amortized methods.
 A more in-depth discussion about amortization can be found in \cref{sec:sup_sa}.
 We propose to exploit the concept of amortization but to apply it at a different granularity, leading to our notion of \textit{plate amortization}.
 
Similar to amortizing across the different data samples $\tX$, we amortize across the different ground RVs $\{\theta_{i,n}\}_{n=0 .. N_i}$ corresponding to the same RV template $\theta_i$.
Instead of casting every flow $\mathcal{F}_{i,n}$, defined in \cref{eq:q_full}, as a separate, fully-parameterized flow, we will share some parameters across the $\mathcal{F}_{i,n}$.
To the template $\theta_i$, we associate a conditional flow $\mathcal{F}_i(\ \cdot \ ; \phi_i, \bullet)$ with weights $\phi_i$ shared across all the $\{\theta_{i,n}\}_{n=0 .. N_i}$.
 The flow $\mathcal{F}_{i,n}$ associated with a given ground RV $\theta_{i,n}$ will be an instance of this conditional flow, conditioned by an encoding $\tE_{i,n}$:
 \begin{equation}
 \label{eq:plate_amo}
     \begin{aligned}
        \mathcal{F}_{i,n} = \mathcal{F}_{i}(\ \cdot \ ; \phi_i, \tE_{i,n})
        \quad \text{yielding} \quad
        q_{i,n} = q_{i,n}(\theta_{i,n} | \pi(\theta_{i,n}) ; \phi_i, \tE_{i,n})
     \end{aligned}
 \end{equation}

 The distributions $q_{i,n}$ thus have 2 sets of weights, $\phi_i$ and $\tE_{i,n}$, creating a parameterization trade-off.
 Concentrating all of $q_{i,n}$'s parameterization into $\phi_i$ results in all the ground RVs $\theta_{i,n}$ having the same posterior distribution.
 On the contrary, concentrating all of $q_{i,n}$'s parameterization into $\tE_{i,n}$ allows the $\theta_{i,n}$ to have completely different posterior distributions.
 But in a large cardinality setting, this freedom can result in a massive number of weights, proportional to the number of ground RVs times the encoding size.
 This double parameterization is therefore efficient when the majority of the weights of $q_{i,n}$ is concentrated into $\phi_i$.
 Using normalizing flows $\mathcal{F}_i$, the burden of approximating the correct parametric form for the posterior is placed onto $\phi_i$, while the $\tE_{i,n}$ encode lightweight summary statistics specific to each $\theta_{i,n}$. For instance, $\mathcal{F}_i$ could learn to model a Gaussian mixture distribution, while the $\tE_{i,n}$ would encode the location and variance of each mode for each ground RV. Encodings $\tE_{i,n}$ allow to individualize for $\theta_{i,n}$ only the strictly necessary information necessary to approximate $\theta_{i,n}$'s posterior. \tmlrrebutal{A more in-depth analysis of the impact of plate amortization on the expressivity of the variational family $\mathcal{Q}$ can be found in \cref{sec:inference_gap}.}

 \paragraph{Intermediate summary}
 This section defined $q$, the variational distribution to approximate full model $\mathcal{M}$'s posterior.
 $q$ features \textit{plate amortization}, which helps maintain a tractable number of weights as the cardinality of $\mathcal{M}$ augments.
 The next section introduces a stochastic scheme to train $q$.
 Critically, combining shared parameterization and stochastic training does not simply result in combining the advantages of both, but features synergies resulting in faster training, as further explained in \cref{sec:sharing_learning}.
\subsection{PAVI stochastic training}
\label{sec:training}
\paragraph{Reduced model}
Our goal is to train the variational distribution $q(\Theta)$, defined in \cref{eq:q_full}, that approximates the posterior $p(\Theta | X)$.
$q$ corresponds to the \textit{full} model $\mathcal{M}$. $\mathcal{M}$ typically features large plate cardinalities $\operatorname{Card}(\mathcal{P})$ ---with many subject groups and subjects per group--- and thus many ground RVs, making it computationally intractable.
We will therefore train $q$ stochastically, over smaller subsets of RVs at a time.
In this section, we interpret stochastic training as the training over a \textit{reduced} model $\redu{\mathcal{M}}$.

Instead of inferring directly over $\mathcal{M}$, we will train over a smaller replica of $\mathcal{M}$.
To this end, we instantiate the template $\mathcal{T}$ into a second HBM $\redu{\mathcal{M}}$, the \textit{reduced} model, of tractable plate cardinalities $\redu{\operatorname{Card}}(\mathcal{P}) \ll \operatorname{Card}(\mathcal{P})$.
$\redu{\mathcal{M}}$ has the same template as $\mathcal{M}$, meaning the same dependency structure and the same parametric form for its distributions.
The only difference lies in $\redu{\mathcal{M}}$'s smaller cardinalities, resulting in fewer ground RVs, as visible in \cref{fig:PAVI_training}.

\paragraph{Reduced distribution and loss}
\label{sec:stochastic_training}
Here we define the distribution $\redu{q}$ used in stochastic training.
$\redu{q}$ features the smaller cardinalities of the reduced model $\redu{\mathcal{M}}$, making it computationally tractable.
At each optimization step $t$, we randomly choose inside $\mathcal{M}$ paths of reduced cardinality, as visible in \cref{fig:PAVI_training}.
Selecting paths is equivalent to selecting from $X$ a subset $\redu{X}[t]$ of size $\redu{N}_X$, and from $\Theta$ a subset $\redu{\Theta}[t]$.
For a given $\theta_i$, we denote as $\mathcal{B}_{i}[t]$ the batch of selected ground RVs, of size $\redu{N}_i$.
$\mathcal{B}_{X}[t]$ equivalently denotes the batch of selected observed RVs.
Inferring over $\redu{\Theta}[t]$, we will simulate training over the full distribution $q$:
\begin{equation}
\label{eq:q_stoc}
    \begin{aligned}
     \log \redu{q}(\redu{\Theta}[t]) &= \sum_{i=1}^I \frac{N_i}{\redu{N}_i} \sum_{\mathclap{\qquad n \in \mathcal{B}_{i}[t]}} \log q_{i,n}(\theta_{i,n} | \pi(\theta_{i,n}))
    \end{aligned}
\end{equation}
where the factor $N_i / \redu{N}_i$ emulates the observation of as many ground RVs as in $\mathcal{M}$ by repeating the RVs from $\redu{\mathcal{M}}$ \citep{SVI}.
Similarly, the loss used at step $t$ is the reduced ELBO constructed using $\redu{X}[t]$ as observed RVs:
\begin{equation}
\label{eq:ELBO_stoc}
    \begin{aligned}
    \redu{\operatorname{ELBO}}[t] &= \mathbb{E}_{\redu{\Theta} \sim \redu{q}}\left[ \log \redu{p}(\redu{X}[t], \redu{\Theta}[t])  - \log \redu{q}(\redu{\Theta}[t]) \right] \\
    \log \redu{p}(\redu{X}[t], \redu{\Theta}[t]) &= \frac{ N_X}{\redu{N}_X} \sum_{\mathclap{\qquad n \in \mathcal{B}_{X}[t]}} \log p_X(x_n | \pi(x_n)) + \sum_{i=1}^I \frac{N_i}{\redu{N}_i} \sum_{\mathclap{\qquad n \in \mathcal{B}_{i}[t]}} \log p_i(\theta_{i,n} | \pi(\theta_{i,n}))
    \end{aligned}
\end{equation}
This scheme can be viewed as the instantiation of $\redu{\mathcal{M}}$ over batches of $\mathcal{M}$'s ground RVs.
In \cref{fig:PAVI_training}, we see that $\redu{q}$ has the cardinalities of $\redu{\mathcal{M}}$ and replicates its conditional dependencies.
This training is analogous to stochastic VI \citep{SVI}, generalized with multiple hierarchies, dependencies in the posterior, and mini-batches of RVs.

\paragraph{Sharing learning across plates}
\label{sec:sharing_learning}
Here we detail how our shared parameterization, detailed in \cref{sec:plate_amortization}, combined with our stochastic training scheme, results in faster inference.
In stochastic VI~\citep[SVI,][]{SVI}, every $\theta_{i,n}$ corresponding to the same template $\theta_i$ is associated with individual weights.
Those weights are trained only when the algorithm visits $\theta_{i,n}$, that is to say, at step $t$ when $n \in \mathcal{B}_{i}[t]$.
As plates become larger, this event becomes rare.
If $\theta_{i,n}$ is furthermore associated with a highly-parameterized density estimator ---such as a normalizing flow--- many optimization steps are required for $q_{i,n}$ to converge.
The combination of those two items leads to slow training and makes inference impractical in contexts such as Neuroimaging, which can feature millions of RVs.
With plate amortization, we aim to unlock inference in those large regimes by reducing the training time. 

Instead of treating the ground RVs $\theta_{i,n}$ independently, we share the learning across plates.
Due to the problem's plate structure, we consider the inference over the $\theta_{i,n}$ as different instances of a common density estimation task.
In PAVI, a large part of the parameterization of the estimators $q_{i,n}(\theta_{i,n} | \pi(\theta_{i,n}); \phi_i, \tE_{i,n})$ is mutualized via the plate-wide-shared weights $\phi_i$.
This means that most of the weights of the flows $\mathcal{F}_{i,n}$, concentrated in $\phi_i$, are trained at every optimization step across all the selected batches $\mathcal{B}_{i}[t]$.
This results in drastically faster convergence than SVI, as seen in experiment \ref{sec:exp_convergence_speed}.

\subsection{Encoding schemes}
\paragraph{PAVI-F and PAVI-E schemes}
\label{sec:encoding_schemes}
PAVI shares the parameterization and learning of density estimators across an HBM's plates. 
In practice the distributions $q_{i,n}(\theta_{i,n} | \pi(\theta_{i,n}); \phi_i, \tE_{i,n})$ from \cref{eq:plate_amo} with different $n$ only differ through the value of the encodings $\tE_{i,n}$.
We detail two schemes to derive those encodings:

\underline{Free plate encodings (PAVI-F)}
In our core implementation, $\tE_{i,n}$ are free weights.
We define encoding arrays with the cardinality of the full model $\mathcal{M}$, one array $\tE_i = [ \tE_{i,n}]_{n=0..N_i}$ per template $\theta_i$.
This means that an additional ground RV ---for instance, adding a subject in a population study--- requires an additional encoding vector.
The associated increment in the total number of weights is much lighter than adding a fully parameterized normalizing flow, as would be the case in the non-plate-amortized regime.
The PAVI-F scheme cannot be sample amortized: when presented with an unseen $\tX$, though $\phi_i$ can be kept as an efficient warm start, the optimal values for the encodings $\tE_{i,n}$ have to be searched again.

During training, the encodings $\tE_{i,n}$ corresponding to $n \in \mathcal{B}_{i}[t]$ are sliced from the arrays $\tE_i$ and are optimized for along with $\phi_i$.
In the toy example from \cref{fig:PAVI_training}, at $t=0$, $\mathcal{B}_{1}[0] = \{ 1, 2 \}$ and the trained encodings are $\{ \tE_{1,1}, \tE_{1,2} \}$, and at $t=1$  $\mathcal{B}_{1}[1] = \{ 0, 1 \}$ and we train $\{ \tE_{1, 0}, \tE_{1,1} \}$.

\underline{Deep set encoder (PAVI-E)}
The parameterization of PAVI-F scales lightly but linearly with $\operatorname{Card}(\mathcal{P})$.
Though lighter than the non-plate-amortized case, this scaling could still become unaffordable in large population studies.
We thus propose an alternate scheme, PAVI-E, with a parameterization independent of cardinalities.
In this more experimental scheme, free encodings are replaced by an encoder $f$ with weights $\eta$ applied to the observed data: $\tE = f(\tX; \eta)$.
As encoder $f$ we use a \textit{deep-set} architecture, detailed in \cref{sec:PAVI-E_details}~\citep{deep_sets, ST, agrawal_amortized_2021}.
Due to the plate structure, the observed $X$ features multiple permutation invariances ---across data points corresponding to i.i.d. RVs.
Deep sets are attention architectures that can model generic permutation invariant functions.
As such, they constitute a natural design choice to incorporate the problem's invariances.
The PAVI-E scheme allows for \textit{sample amortization} across different data samples $\tX_0, \tX_1, ...$, as described in \cref{sec:plate_amortization}.
Note that an encoder will be used to generate the encodings whether the inference is sample amortized or not.

During training, shared learning is further amplified as all the architecture's weights ---$\phi_i$ and $\eta$--- are trained at every step $t$.
To collect the encodings to plug into $\redu{q}$, we build up on a property of $f$: \textit{set size generalization} \citep{deep_sets}.
Instead of encoding the full-sized data $\tX$, $f$ is applied to the slice $\redu{\tX}[t]$.
This amounts to aggregating summary statistics across a subset of the observed data instead of the full data~\citep{ST, agrawal_amortized_2021}.
The PAVI-E scheme has an even greater potential in the sample amortized context: we train a sample amortized family over the lightweight model $\redu{\mathcal{M}}$ and use it "for free" to infer over the heavyweight model $\mathcal{M}$.

\paragraph{Stochastic training and bias}

A key consideration is a potential bias introduced by our stochastic scheme.
Namely, training stochastically over a variational family $\mathcal{Q}$, we want to converge to the same solution $q^*$ as if we trained over the entirety of $\mathcal{Q}$.
In this section, we show that the PAVI-F scheme is unbiased.
In contrast, the PAVI-E scheme is theoretically biased ---though we seldom noticed any negative impact of that bias in practice.
A more thorough analysis can be found in \cref{sec:no_bias}.
\tmlrrebutal{Note that in this manuscript the term \textit{bias} systematically refers to the stochastic training scheme.
A different form of bias consists in the limited expressivity of the variational family $\mathcal{Q}$ which may not contain the true posterior $p(\Theta | X)$.
We refer to this other bias as the variational family's \textit{gap}, as further detailed in \cref{sec:inference_gap}.}

To show that our training scheme is unbiased, we need to prove that the expectation of our stochastic loss is equal to the non-stochastic loss.
This amounts to showing that:
\begin{equation}
\label{eq:no_bias_main}
    \begin{aligned}
    \mathbb{E}_\text{paths} \left[ \redu{\operatorname{ELBO}}[t] \right]
        &= \operatorname{ELBO} \\
        &= \mathbb{E}_{\Theta \sim q}\left[ \log p(X, \Theta) - \log q(\Theta) \right]
    \end{aligned}
\end{equation}
where $\mathbb{E}_\text{paths}$ loosely refers to the paths that are sampled stochastically into the full model $\mathcal{M}$'s graph, as defined in \cref{sec:stochastic_training}.
\Cref{eq:no_bias_main} means that the expectation of the reduced ELBO over all the stochastic paths that can be sampled inside the full model's graph is equal to the full ELBO.
To prove \cref{eq:no_bias_main}, it is sufficient to prove that:
\begin{equation}
\label{eq:no_bias_q_main}
    \mathbb{E}_\text{paths} \left[ \log \redu{q}(\redu{\Theta}[t]) \right]
        = \log q(\Theta)
\end{equation}
meaning that the expectation of the reduced distribution $\redu{q}$ over the stochastic paths equals the full distribution $q$.
In \cref{sec:no_bias} we show that:
\begin{equation}
\label{eq:E_q_redu}
    \begin{aligned}
     \mathbb{E}_\text{paths}  \left[ \log \redu{q}(\redu{\Theta}[t]) \right]
        &= \sum_{i=1}^I \sum_{n=0}^{N_i} \mathbb{E}_\text{paths} \left[\log q_{i,n}(\theta_{i,n} | \pi(\theta_{i,n}); \tE_{i,n})\right]
    \end{aligned}
\end{equation}
an expression that mimics the definition of $q$ in \cref{eq:q_full} ---albeit the expectations over paths.
The rest of the bias analysis depends on the encoding scheme.

\underline{Free plate encodings (PAVI-F)}
In the PAVI-F scheme, the encodings $\tE_{i,n}$ are free weights.
As a consequence, their value does not depend on the paths that are selected by the stochastic algorithm.
This means that $\mathbb{E}_\text{paths} \left[\log q_{i,n}(\theta_{i,n} ; \tE_{i,n})\right] = \log q_{i,n}(\theta_{i,n} ; \tE_{i,n})$, proving \cref{eq:no_bias_q_main} and the unbiasedness of the PAVI-F scheme.
With the PAVI-F scheme, training over $\redu{\mathcal{M}}$, we converge to the same distribution as if training directly over $\mathcal{M}$.

\underline{Deep set encoder (PAVI-E)}
In the PAVI-E scheme, encodings are obtained by applying the encoder $f$ to a slice of the observed data.
With the implementation presented in this work, the value of $\tE_{i,n}$ will depend on which children of $\theta_{i,n}$ the stochastic algorithm selects.
For instance, the encoding for a group will depend on which subjects are selected inside the group to compute the group's summary statistics.
Though computationally efficient, our implementation is thus theoretically biased.
The negative impact of this bias on PAVI-E's performance was seldom noticeable and always marginal throughout our experiments.

\paragraph{Technical summary}
In \cref{sec:archi}, we derived an architecture sharing its parameterization across a model's plates.
This allows performing inference in large cardinality regimes without incurring an exploding number of weights.
In \cref{sec:training} we derived a stochastic scheme to train this architecture over batches of data.
Stochastic training allows large-scale inference without incurring an exploding memory and computing.
Off-the-shelf stochastic VI would however result in significantly slower inference.
Our novelty lies in our original combination of amortization and stochastic training, which avoids this slow-down, as demonstrated in the following experiments.

\section{Results and discussion}
\label{sec:exps}

In this section, we show how PAVI unlocks hierarchical Bayesian model inference for large-scale problems by matching the inference quality of SOTA methods while providing faster convergence and lighter parameterization.
Our experiments also highlight the differences between our two encoding schemes PAVI-E and PAVI-F.
In summary:
\begin{enumerate}
    \item \Cref{sec:exp_convergence_speed} shows how plate amortization results in faster convergence compared to non-plate-amortized SVI;
    \item \Cref{sec:exp_encoding_size} illustrates the role of the encodings $\tE_{i,n}$ as summary statistics in inference;
    \item \Cref{sec:exp_scaling} exemplifies --against baselines-- PAVI's favorable scaling as a problem's cardinality augments;
    \item \Cref{sec:exp_fMRI} showcases the practical utility of PAVI by applying the method to a challenging population study.
\end{enumerate}

\paragraph{Supplemental experiments}
In \cref{sec:exp_card_redu}, we evaluate the impact of the reduced model cardinalities on performance.
\tmlrrebutal{
In \cref{sec:sup_comparisons}, we compare our method against baselines over a mixture model; over a model featuring the aggregation of higher-order summary statistics; and over a smaller version of our Neuroimaging model used in \cref{sec:exp_fMRI}.
Those models complement the analysis of our main text, which is mostly concentrated on Gaussian models.
Gaussian models constitute a standard model featuring an approximate closed-form posterior which we use for validation.
}

\paragraph{ELBO metric}
Throughout this section, we use the ELBO as a proxy for the KL divergence between the variational posterior and the unknown true posterior~\citep{blei_variational_2017}.
ELBO is measured across 20 samples $\tX$, with 5 repetitions per sample.
The ELBO allows us to compare the relative performance of different architectures on a given inference problem.
In \cref{sec:sanity}, we provide sanity checks to assess the quality of the results.

\begin{figure}
    \centering
    \begin{subfigure}[t]{0.47\textwidth}
        \centering
        \includegraphics[width=\textwidth]{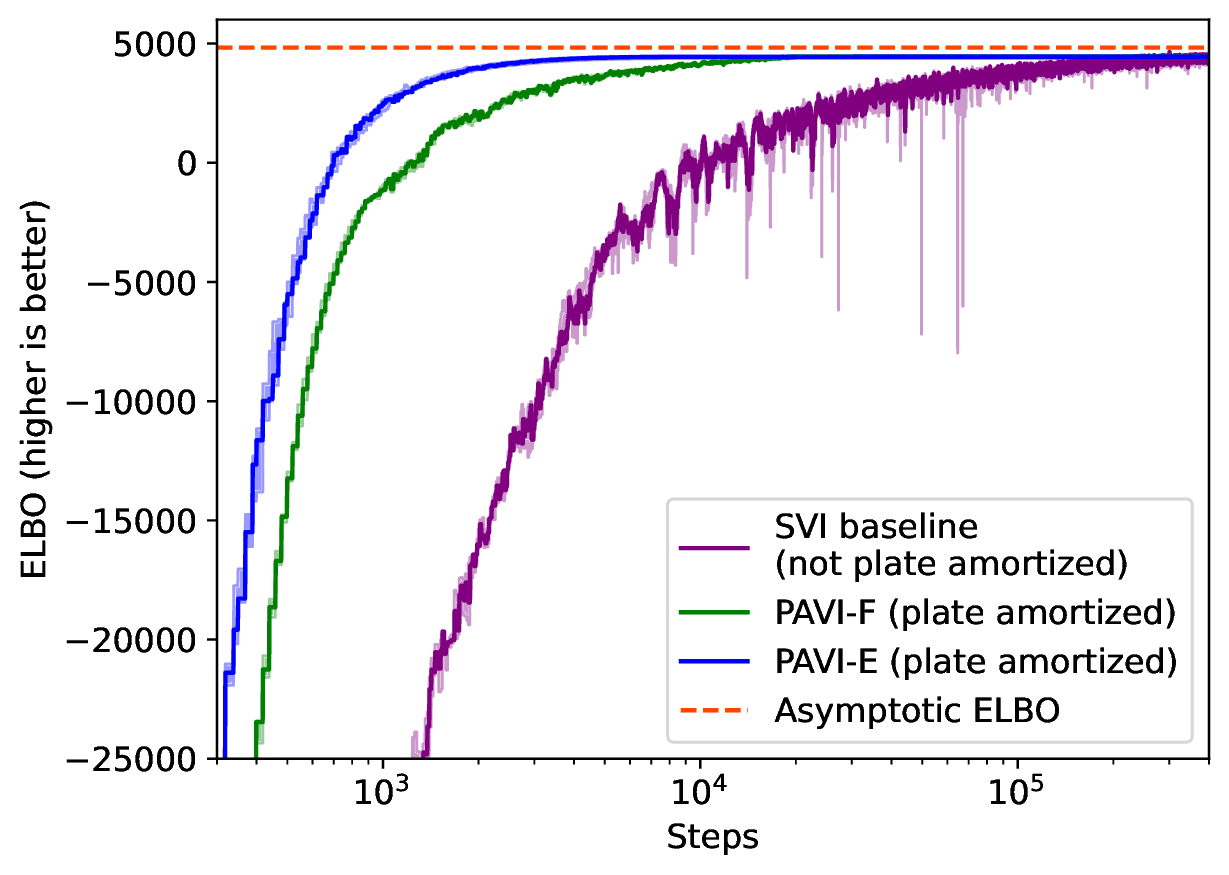}
    \end{subfigure}
    \hfill
    \begin{subfigure}[t]{0.48\textwidth}
        \centering
        \includegraphics[width=\textwidth]{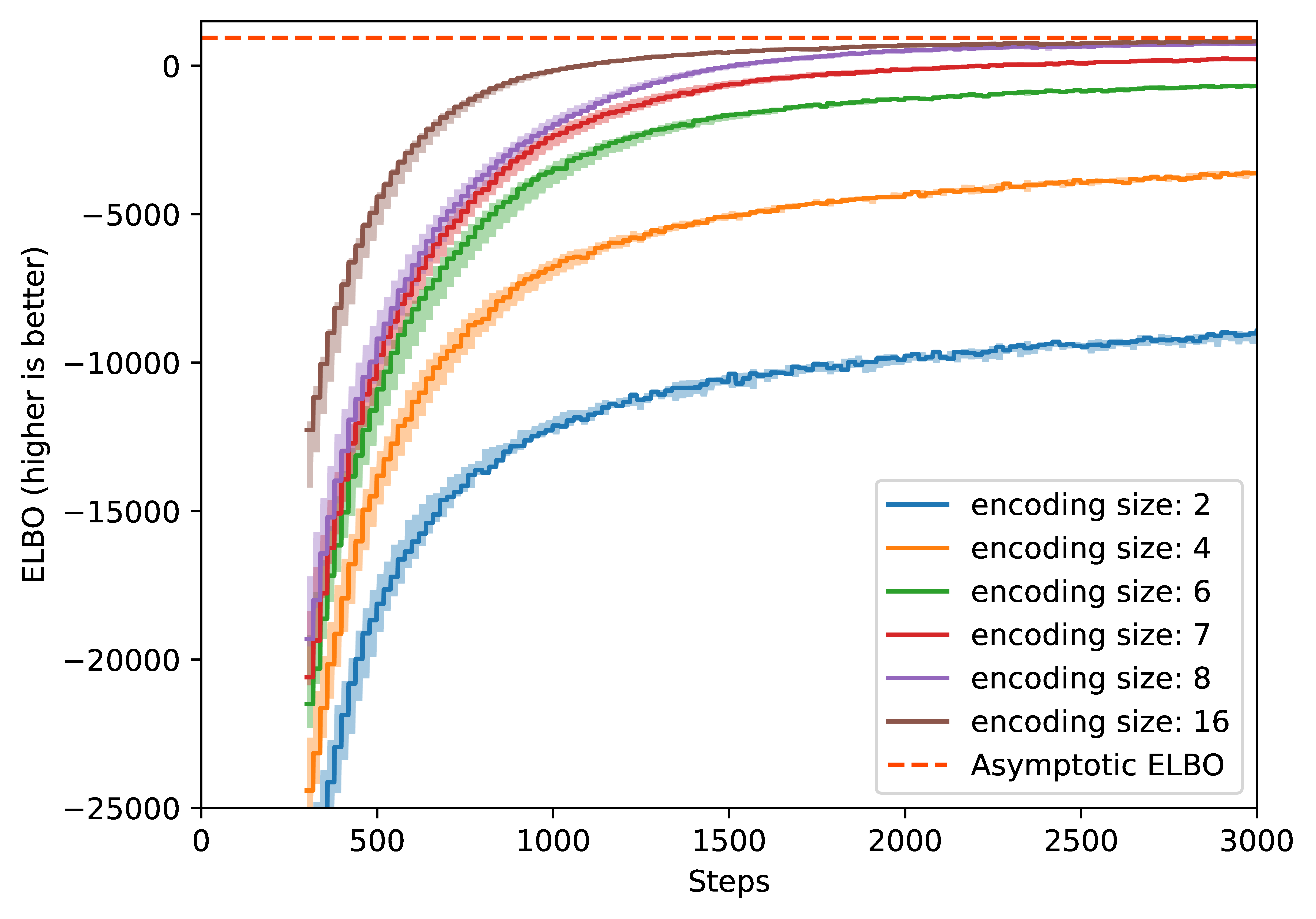}
    \end{subfigure}
    \vspace{-10pt}
    \caption{\textbf{Left panel: Plate amortization increases convergence speed} Plot of the ELBO (higher is better) as a function of the optimization steps (log-scale) for our methods PAVI-F (in green) and PAVI-E (in blue) versus a non-plate-amortized baseline (in purple).
    \tmlrrebutal{Standard deviation across repetitions is displayed as a shaded area.
    A dashed line denotes the asymptotic closed-form performance.}
    Due to plate amortization, our method converges ten to a hundred times faster to the same asymptotic ELBO as its non-plate-amortized counterpart.;
    \textbf{Right panel: Encodings as ground RVs summary statistics} Plot of the ELBO (higher is better) as a function of the optimization steps for the PAVI-F architecture with increasing encoding sizes.
    \tmlrrebutal{Standard deviation across repetitions is displayed as a shaded area.
    A dashed line denotes the asymptotic closed-form performance.}
    As the encoding size augments, so does the asymptotic performance until reaching the dimensionality of the posterior's sufficient statistics ($D=8$), after which performance plateaus.
    Encoding size allows for a clear trade-off between memory and inference quality.}
    \label{fig:convergence_speed}
    \label{fig:encoding_size}
    \vspace{-10pt}
\end{figure}

\tmlrrebutal{
\paragraph{Gaussian random effects (GRE) model}
In our experiments \cref{sec:exp_convergence_speed}, \ref{sec:exp_encoding_size}, \ref{sec:exp_scaling}, we focus on a standard hierarchical model: a Gaussian random effects model (GRE)\citep{diggle_analysis_2013, Gelman_book}.
The GRE model constitutes an intuitive inference problem ---inferring group and population means given observations--- and features an approximate closed-form solution, as described in the next paragraph.
This closed-form solution helps us display theoretical bounds for the ELBO, for instance in \cref{fig:convergence_speed}.

The GRE model can be described using the following set of equations: 
\begin{equation}
\label{eq:GRE}
\begin{aligned}
    X_{n_1,n_0} |\theta_{1,n_1} &\sim \mathcal{N}(\theta_{1,n_1}, \sigma_x^2)
    \quad
    \substack{\forall n_1=1..\operatorname{Card}(\mathcal{P}_1) \\ \forall n_0=1..\operatorname{Card}(\mathcal{P}_0)}
    \\
    \theta_{1,n_1} | \theta_{2,0} &\sim \mathcal{N}(\theta_{2,0}, \sigma_1^2)
    \quad \ \ 
    \substack{\forall n_1=1..\operatorname{Card}(\mathcal{P}_1) \\}
    \qquad
    \theta_{2,0} &\sim \mathcal{N}(\vec 0_D, \sigma_2^2) \enspace ,
\end{aligned}
\end{equation}
where $D$ represents the data $\tX$'s feature size, with group means $\theta_1$ and population means $\theta_2$ as D-dimensional Gaussians.
The GRE model features two nested plates: the group plate $\mathcal{P}_1$ and the sample plate $\mathcal{P}_0$ as in \cref{fig:PAVI_principle}.
Taking our introductory example from \cref{sec:HBMs}, $X_{n_1,n_0}$ represents the weight for subject $n_0$ in group $n_1$.
$\theta_{1,n_1}$ represents the mean weight in the group $n_1$.
$\theta_{2,0}$ represents the mean weight in the population.
Inferring over the GRE model, the objective is to retrieve the posterior distribution of the group and population means given the observed sample.
}

\tmlrrebutal{
\paragraph{Asymptotic closed-form ELBO}
As a baseline for comparison, we compare the ELBO of various methods to an approximate closed-form baseline's ELBO.
Though a closed-form posterior cannot be derived in the 3-level case, a good approximation can be constructed using Gaussian distributions centered on the empirical group and population means.
This asymptotic ELBO is represented using dashed lines in \cref{fig:convergence_speed} and \ref{fig:scaling}.
}

\subsection{Plate amortization and convergence speed}
\label{sec:exp_convergence_speed}
In this experiment, we illustrate how plate amortization results in faster training.

We use the GRE model, described in \cref{eq:GRE}, setting $D=8$, $\operatorname{Card}(\mathcal{P}_1)=100$ and $\redu{\operatorname{Card}}(\mathcal{P}_1)=2$.
\tmlrrebutal{
In this experiment, we set $\redu{\operatorname{Card}}(\mathcal{P}_1) \ll \operatorname{Card}(\mathcal{P}_1)$.
This emulates a regime in which SVI is slow because only a small fraction ---of size $\redu{\operatorname{Card}}(\mathcal{P}_1)$--- of a large parameter space ---of size $\operatorname{Card}(\mathcal{P}_1)$--- gets optimized at a given stochastic training step.}
We compare our PAVI architecture to a baseline with the same architecture, trained stochastically with SVI~\citep{SVI}, but without plate amortization.
The only difference is that ground RVs $\theta_{i,n}$ are associated in the baseline to individual fully-parameterized flows $\mathcal{F}_{i,n}$ instead of sharing the same conditional flow $\mathcal{F}_i$, as further described in \cref{sec:plate_amortization}.

\Cref{fig:convergence_speed} (left) displays the evolution of the ELBO across training steps for the baseline and PAVI with free encoding (PAVI-F) and deep-set encoders (PAVI-E).
We see that both plate-amortized methods reach asymptotic ELBO equal to the non-plate-amortized baseline's but with orders of magnitudes faster convergence and more numerical stability.
This stems from the individual flows $\mathcal{F}_{i,n}$ in the baseline only being trained when the stochastic algorithm visits the corresponding $\theta_{i,n}$.
In contrast, our shared flow $\mathcal{F}_i$ is updated at every optimization step in PAVI.
Intuitively, the PAVI-E scheme should converge faster than PAVI-F by sharing the training not only of the conditional flows but also of the encoder across the different optimization steps.
However, the computation required to derive the encodings from the observed data results in longer optimization steps and in slower inference, as illustrated in \cref{sec:exp_scaling}.
The usage of an encoder nonetheless allows for \textit{sample amortization} with the PAVI-E scheme, which is impossible with the PAVI-F scheme.

\subsection{Impact of encoding size}
\label{sec:exp_encoding_size}
Here we illustrate the role of encodings as ground RV posterior's summary statistics, as further described in \cref{sec:plate_amortization}.
We use the GRE HBM detailed in \cref{eq:GRE}, using $D=8$, $\operatorname{Card}(\mathcal{P}_1)=20$ and $\redu{\operatorname{Card}}(\mathcal{P}_1)=2$.
We use a single PAVI-F architecture, varying the size of the encodings $\tE_{i,n}$, which are defined in \cref{sec:encoding_schemes}.

Due to plate amortization, encodings determine how much individual information each RV $\theta_{i,n}$'s posterior is associated with.
The encoding size ---varying from $2$ to $16$--- is to be compared with the problem's dimensionality, $D=8$.
In GRE, $D=8$ corresponds to the size of the sufficient statistics needed to reconstruct the posterior of a group mean since all other statistics, such as the variance, are shared between the group means.

\Cref{fig:encoding_size} (right) shows how the asymptotic performance steadily increases when the encoding size augments and plateaus once reaching the sufficient summary statistic size $D=8$.
Interestingly, increasing the encoding size also leads to faster convergence: redundancy can likely be exploited in the optimization.
Increasing the encoding size also leads experimentally to diminishing returns in terms of performance.
This property can be exploited in large settings to drastically reduce the memory footprint of inference while maintaining acceptable performance by choosing the encoding size approximately equal to the expected size of the sufficient statistics.
Encoding size appears as an unequivocal hyperparameter allowing to trade inference quality for computational efficiency.

\subsection{Scaling with plate cardinalities}
\begin{figure}[t]
    \centering
    \includegraphics[width=0.95\textwidth]{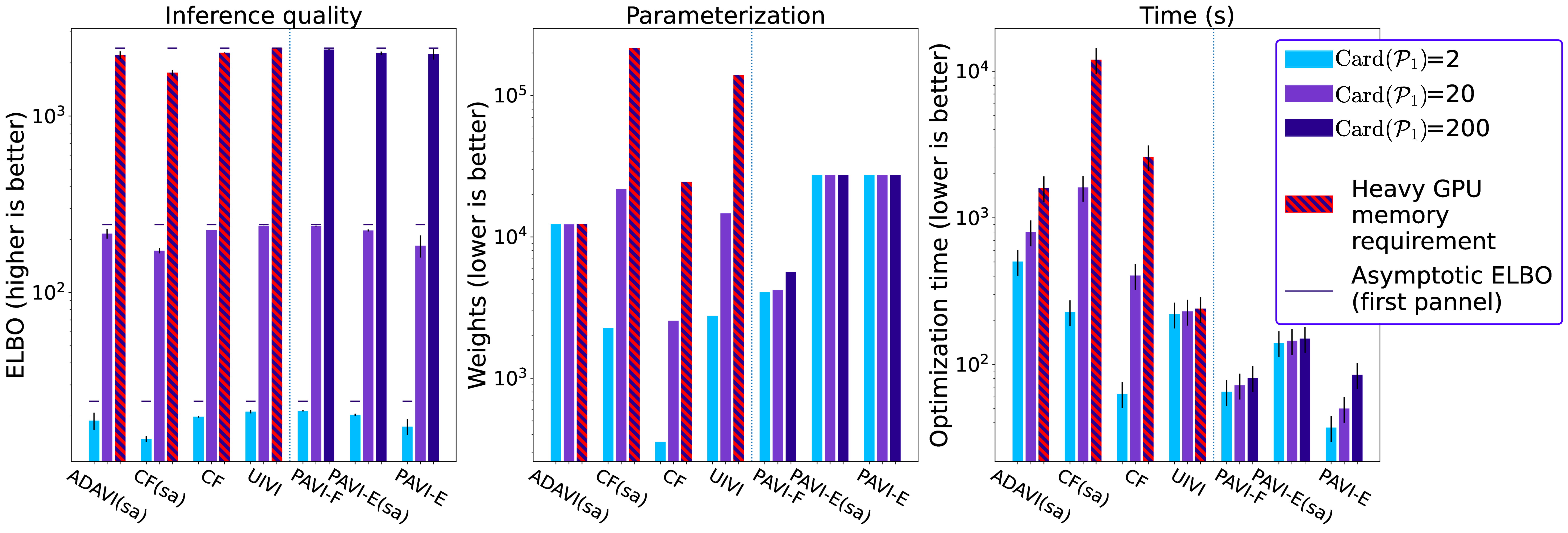}
    \vspace{-8pt}
    \caption{
    \rebutal{\textbf{PAVI provides favorable parameterization and training time as the cardinality of the target model augments}}
    Baselines are compared in each panel, with the suffix (sa) indicating \textit{sample amortization}.  
    Our architecture PAVI is displayed on the right of each panel.
    We augment the cardinality $\operatorname{Card}(\mathcal{P}_1)$ of the GRE model, which is described in \cref{eq:GRE}.
    While doing so, we compare three different metrics.
    \textit{In the first panel:} inference quality, as measured by the ELBO.
    \tmlrrebutal{An asymptotic closed-form ELBO is displayed using a dark blue dash.}
    None of the presented SOTA architecture's performance degrades as the cardinality of the problem augments.
    \textit{In the second panel:} parameterization, comparing the number of trainable weights of each architecture.
    PAVI --similar to ADAVI-- displays a constant number of weights as the cardinality of the problem increases ---or almost constant for PAVI-F.
    \textit{Third panel:} GPU training time.
    Benefiting from learning across plates, PAVI has a short and almost constant training time as the cardinality of the problem augments.
    At $\operatorname{Card}(\mathcal{P}_1)=200$, CF, UIVI, and ADAVI required large GPU memory, a constraint absent from PAVI due to its stochastic training.
    \tmlrrebutal{For convenience, the results used to plot this figure are reproduced in supplemental \cref{tab:scaling}}.}
    \label{fig:scaling}
\end{figure}
\label{sec:exp_scaling}
Here we put in perspective the gains from plate amortization when scaling up an inference problem's cardinality.
We consider the GRE model in \cref{eq:GRE} with $D=2$ and augment the plate cardinalities $(\operatorname{Card}(\mathcal{P}_1),\redu{\operatorname{Card}}(\mathcal{P}_1)):\ (2, 1) \rightarrow (20, 5) \rightarrow (200, 20)$.
In doing so, we augment the number of parameters $\Theta:\ 6 \rightarrow 42 \rightarrow 402$.

\textbf{Baselines}
We compare our PAVI architecture against three state-of-the-art automatic VI baselines.
\begin{itemize}
    \item \textbf{Cascading Flows (CF)} \citep{CF} is a non-plate-amortized structured VI architecture improving on the baseline presented in \cref{sec:exp_convergence_speed}.
    CF pushes the prior $p$ into the posterior $q$ using \textit{Highway Flows}.
    CF follows a cascading dependency structure complemented by a backward auxiliary coupling.
    CF thus consists in a structured baseline that does not pay particular attention to scalability to large cardinalities;
    \item \textbf{Automatic Dual Amortized VI (ADAVI)} \citep{ADAVI} is a structured VI architecture with constant parameterization with respect to a problem's cardinality but large training times and memory.
    ADAVI has several limitations compared to PAVI.
    First, ADAVI implements a Mean Field approximation \citep{blei_variational_2017} while PAVI implements a cascading flow.
    This means that, contrary to PAVI, ADAVI can be biased in cases where the posterior features strong dependencies.
    Second, ADAVI is limited to pyramidal HBMs, while PAVI tackles generic plate-enriched HBMs, making it more general.
    Third, ADAVI is limited to a full-model sample-amortized variant, which ultimately caps the cardinality of problems it can be used on.
    ADAVI thus consists in a structured baseline that tackles the parameterization hurdle of large cardinalities, but not the computational efficiency hurdle;
    \item \textbf{Unbiased Implicit VI (UIVI)} is an unstructured implicit VI architecture.
    UIVI infers over the full parameter space $\Theta$, without any SVI-amenable factorization ---contrary to CF, ADAVI, and PAVI.
    To do so, UIVI reparameterizes a base distribution with a stochastic transform.
    UIVI does not explicitly define a density $q$ and relies on optimization steps intertwined with MCMC runs.
    UIVI thus consists in a non-structured VI baseline that does not pay particular attention to scalability to a large parameter space
    This means that UIVI could not be applied above a certain cardinality due to its impossibility to be stochastically trained.
\end{itemize}
For all architectures, we indicate with the suffix \textit{(sa) sample amortization}, as defined in \cref{sec:plate_amortization}.
More implementation details can be found in \cref{sec:details_arch}.

As the cardinality of the problem augments, \cref{fig:scaling} shows how PAVI maintains a state-of-the-art inference quality while being more computationally attractive.

\textbf{Parameterization}
In terms of parameterization, both ADAVI and PAVI-E provide a heavyweight but constant parameterization as the cardinality $\operatorname{Card}(\mathcal{P}_1)$ of the problem augments.
This is due to both methods' usage of an encoder of the observed data, which makes their parameterization independent of the problem's cardinality.
Comparatively, both CF and PAVI-F's parameterization scale linearly with $\operatorname{Card}(\mathcal{P}_1)$, but with a drastically lighter augmentation for PAVI-F.
The difference can be explained by the additional weights required by each architecture for an additional ground RV.
CF requires an additional fully parameterized normalizing flow, whereas PAVI-F only requires an additional lightweight encoding vector.
In detail, PAVI-F's parameterization due to the plate-wide-shared $\phi_1$ represents a constant $\approx 2k$ weights, while the part due to the encodings $\tE_{1,n}$ grows linearly from $16$ to $160$ to $1.6k$ weights.
UIVI's parameterization scales quadratically with the size of the parameter space $\Theta$.
This is due to UIVI's usage of a neural network to regress the weights of a transform applied to a base distribution with the size of $\Theta$.
As the cardinality augments, UIVI's quadratic weight scaling would be limiting before CF and PAVI-F's linear scaling, which would be limiting before ADAVI and PAVI-E's constant scaling.

\textbf{Memory}
Regarding computational budget, PAVI's stochastic training allows for controlled GPU memory during optimization.
This removes the need for a larger memory as the cardinality of the problem augments, a hardware constraint that can become unaffordable at very large cardinalities.
CF could be trained stochastically to remove this memory constraint but, without plate amortization, would suffer from slower inference, as illustrated in \cref{sec:exp_convergence_speed}.
In contrast, UIVI could not be trained stochastically, as it infers over the full parameter space $\Theta$ at once instead of factorizing it.
As a result, UIVI would be ultimately limited by memory to infer over larger problems.

\textbf{Speed}
Regarding convergence speed, PAVI benefits from plate amortization to have orders of magnitude faster convergence compared to structured VI baselines CF and ADAVI.
This means that a stochastically-trained architecture (PAVI) trains faster than non-stochastic baselines (CF, ADAVI)
This result is opposite to the result we would have obtained without plate amortization since SVI slows down inference.
For UIVI, as the cardinality augments, training amounts to evaluating a neural network with increasingly larger layers.
GPU training time is thus constant, but this property would not translate to larger problems as the GPU memory would become insufficient.
Plate amortization is particularly significant for the PAVI-E(sa) scheme, in which a sample-amortized variational family is trained over a dataset of reduced cardinality yet performs "for free" inference over an HBM of  large cardinality.
Maintaining $\redu{\operatorname{Card}}(\mathcal{P}_1)$ constant while $\operatorname{Card}(\mathcal{P}_1)$ augments allows for a constant parameterization \textit{and training time} as the cardinality of the problem augments
This property is not limited to any maximum cardinality, contrary to UIVI.
This is a novel result with strong future potential.
The effect of plate amortization is particularly noticeable at $\operatorname{Card}(\mathcal{P}_1)=200$ between the PAVI-E(sa) and CF(sa) architectures, where PAVI performs sample-amortized inference with $10\times$ fewer weights and $100\times$ lower training time.

Scaling even higher the cardinality of the problem ---$\operatorname{Card}(\mathcal{P}_1)=2000$ for instance--- renders ADAVI, CF and UIVI computationally intractable
In contrast, PAVI maintains a light memory footprint and a short training time, as exemplified in the next experiment.

\tmlrrebutal{
\textbf{Using PAVI on small scale problems}
By design, PAVI is meant to tackle large cardinality problems effectively.
The use of PAVI on smaller problems is thus an open question.
As \cref{fig:scaling} underlines, in the $\operatorname{Card}(\mathcal{P}_1)=2$ case, PAVI-F matches the ELBO and training speed of CF but features a heavier parameterization.
In addition, plate amortizing a variational family theoretically reduces its expressivity, as further explained in \cref{sec:inference_gap}.
In theory, a simpler, non-plate-amortized baseline such as CF could thus be preferred for small problems.
However, in practice, VI amounts to an optimization problem, and the most expressive variational family will not necessarily yield the best performance on a complex problem~\citep{bottou_tradeoffs_2007}.
Our comparative experiments in \cref{sec:sup_comparisons} exemplify this.
In \cref{sec:sup_comparisons}, PAVI's repeated parameterization oftentimes eases the optimization and yields a better ELBO.
We leave for future work the analysis of the impact of PAVI's repeated parameterization ---which could be considered as an inductive bias--- on properties such as inference calibration~\citep{talts_validating_2020}.
In conclusion, when its parameterization is affordable, PAVI remains a relevant choice for small plate-enriched problems.
}

\subsection{Neuroimaging application: full cortex probabilistic parcellation for a cohort of 1,000 subjects}
\label{sec:exp_fMRI}
To illustrate its usefulness, we apply PAVI to a challenging population study for full-cortex functional \textit{parcellation}.

\begin{figure}
    \centering
    \begin{subfigure}[]{0.48\textwidth}
        \centering
        \includegraphics[width=\textwidth]{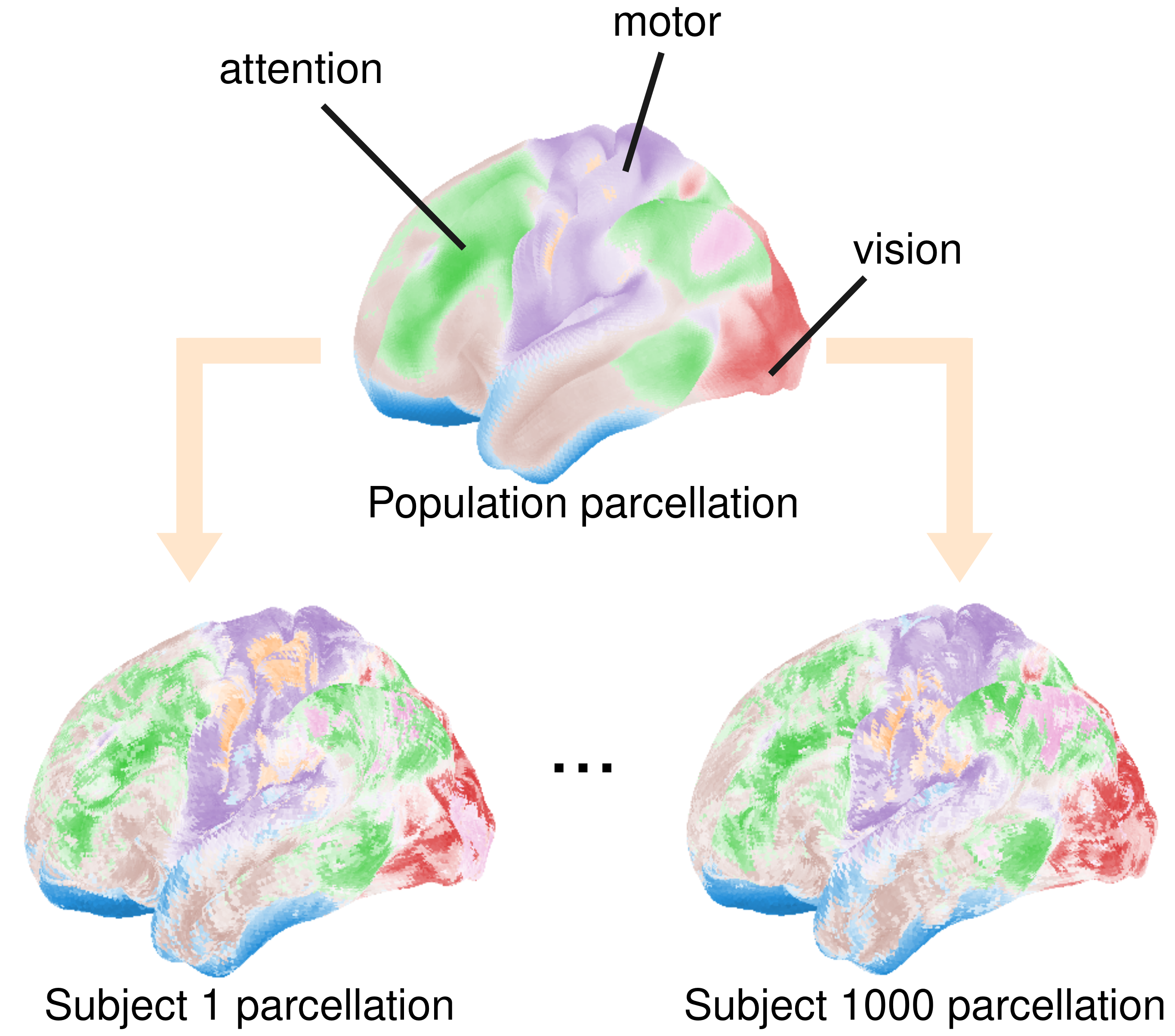}
    \end{subfigure}
    \begin{subfigure}[]{0.48\textwidth}
        \centering
        \begin{tabular}{rl}
             & Accuracy for \\
             & 13 cognitive measures \\
             Method & (Higher is better) \\
            \\ \hline \\
            MS-HBM & $0.1321$ ($\pm 0.0053$) \\
            from \citet{kong_spatial_2018} & \\
            YeoBackProject & $0.1057$ ($\pm 0.0060$) \\
            from \citet{calhoun_multisubject_2012} & \\
            Gordon2017 & $0.0545$ ($\pm 0.0062$) \\
            from \citet{parc_2} & \\
            Wang2015 & $0.1202$ ($\pm 0.0054$) \\
            from \citet{parc_3} & \\
            \textbf{Ours (PAVI)} & $\mathbf{0.1645}$ ($\pm 0.0047$)
        \end{tabular}
    \end{subfigure}
    \caption{\textbf{Probabilistic full cortex parcellation}
    PAVI can be applied in the challenging context of Neuroimaging population studies.
    For a cohort of $1000$ subjects, 2 of which are represented here (in the bottom 2 items) we cluster $60,000$ cortex vertices according to their connectivity with the rest of the brain.
    On the left, we show the obtained probabilistic \textit{parcellations}.
    Each color in the parcellation corresponds to one of 7 functional \textit{network}~\citep{parc_1}.
    Networks represent groups of neurons that co-activate in the brain and can be associated with certain cognitive functions, such as vision or motor control.
    Through our method, we recover each subject's parcellation (at the bottom), which are i.i.d. perturbations of the population's parcellation (at the top).
    Our method also models uncertainty: coloring represents the dominant label for each vertex, and the level of white increases with the uncertainty in the labeling.
    We can use the subject parcellations as features for a cognitive score prediction task, as visible in the table on the right.
    The table shows the mean predictive accuracy across 13 cognitive measures, including memory, pronunciation, processing speed or spatial orientation.
    The baseline methods scores are reproduced from \citet{kong_spatial_2018}'s implementation.
    Our method produces individual maps that are predictive of the subject's cognitive ability.}
    \label{fig:full}
\end{figure}

A \textit{parcellation} clusters brain vertices into different \textit{connectivity networks}: fingerprints describing co-activation with the rest of the brain, as measured using functional Magnetic Resonance Imaging (fMRI).
A parcellation is essentially a clustering of a subject's cortex vertices based on their connectivity with the rest of the brain.
Different subjects exhibit a strong variability, as visible in \cref{fig:full}.
However, fMRI is costly to acquire: few noisy measurements are usually made for a given subject.
It is thus essential to combine information across the different measurements for a given subject and across subjects and to display uncertainty in the results.
Those 2 points motivate hierarchical Bayesian modelling and VI in Neuroimaging~\citep{kong_spatial_2018}: we search the posteriors of connectivity networks and vertex labels, measuring fMRI over a large cohort of subjects.

We use the HCP dataset \citep{HCP}: $2$ acquisitions from a $1000$ subjects, with millions of measures per acquisition, and \textit{over 400 million parameters} $\Theta$ to infer.
We use a model with three plates: subjects, sessions and brain vertices.
We cluster the brain vertices into 7 clusters, following \citet{parc_1}.
None of the baselines presented in \cref{sec:exp_scaling} ---CF, ADAVI, UIVI--- can computationally tackle this high cardinality problem.
We nevertheless show superior performance over those baselines over a tractable problem size with 2,000 parameters in \cref{sec:exp_toy_HCPL}.

Despite the massive dimensionality of the problem, thanks to plate amortization, PAVI converges in less than 5 hours of GPU time.
Results are visible in \cref{fig:full}.
Qualitatively, the recovered networks match existing expert knowledge on the brain's functional organization.
For instance, we recover networks responsible for motor control or vision~\citep{heimEffectiveConnectivityLeft2009, zhangConnectingConceptsBrain2020}.
Quantitatively, using the individual subject's parcellation as feature for a cognitive score regression task, we obtain better scores compared to state-of-the-art methods~\citep{parc_1, parc_2, parc_3, kong_spatial_2018}.
This means that a subject's brain spatial organization ---where each function is "located" in the brain--- can be used to predict the subject's cognitive ability, such as memory, reading, or spatial orientation.

Using PAVI, we obtain state-of-the-art functional parcellations for a thousand subjects, capturing the uncertainty in this challenging task via VI while maintaining a manageable training time despite a 400-million-sized parameter space.

\subsection{\textbf{Conclusion}}
In this work we present the novel PAVI architecture, combining a structured variational family and a stochastic training scheme.
PAVI is based the on concept of plate amortization, allowing to share parameterization and learning across a model's plates.
We demonstrated the positive impact of plate amortization on training speed and scaling to large plate cardinality regimes, making a significant step towards scalable, expressive Variational Inference.

\subsubsection*{Reproducibility statement}
All experiments were performed in Python using the Tensorflow Probability library \citep{dillon_tensorflow_2017}.
All experiments were conducted on computational cluster nodes equipped with a Tesla V100-16Gb GPU and 4 AMD EPYC 7742 64-Core processors.
VRAM intensive experiments in \cref{fig:scaling} were performed on an Ampere 100 PCIE-40Gb GPU.
\Cref{sec:details_arch} lists implementation details for all our synthetic experiments.
\Cref{sec:details_neuro} is related to our Neuroimaging experiment \ref{sec:exp_fMRI}, and details both our data pre-processing steps and our implementation.
As part of our submission we furthermore packaged and release the code associated to our experiments.
\subsubsection*{Broader Impact Statement}
Our Neuroimaging data come from the Human Connectome Project dataset \citep{HCP}.
All data in the HCP is strongly anonymized, as per the \href{https://www.humanconnectome.org/study/hcp-young-adult/project-protocols}{HCP protocols}.
We used in this paper only Open Access imaging data data, following the \href{https://www.humanconnectome.org/study/hcp-young-adult/data-use-terms}{HCP data use terms}.
\subsubsection*{Acknowledgments}
This work was supported by the ERC-StG NeuroLang ID:757672.

This work was performed using HPC resources from GENCI–IDRIS (Grant 2022-102043).

\bibliography{references}
\bibliographystyle{tmlr}

\newpage
\appendix
\section*{Appendix}

\setcounter{figure}{0}
\setcounter{equation}{0}
\renewcommand\thefigure{\thesection.\arabic{figure}}
\renewcommand\thetable{\thesection.\arabic{table}}
\renewcommand\theequation{\thesection.\arabic{equation}}

\section{Supplemental methods}
\subsection{PAVI implementation details}
\subsubsection{Plate branchings and stochastic training}
\label{sec:plate_branching}
As exposed in \cref{sec:stochastic_training}, at each optimization step $t$ we randomly select branchings inside the full model $\mathcal{M}$, branchings over which we \textit{instantiate} the reduced model $\redu{\mathcal{M}}$.
In doing so, we define batches $\mathcal{B}_{i}[t]$ for the RV templates $\theta_i$.
Those batches have to be coherent with one another: they have to respect the conditional dependencies of the original model $\mathcal{M}$.
As an example, if a ground RV is selected as part of $\redu{\mathcal{M}}$, then its parent RV needs to be selected as well.
To ensure this, during the stochastic training we do not sample RVs directly but plates:
\begin{enumerate}
    \item For every plate $\mathcal{P}_p$, we sample without replacement $\redu{\operatorname{Card}}(\mathcal{P}_p)$ indices amongst the $\operatorname{Card}(\mathcal{P}_p)$ possible indices.
    \item Then, for every RV template $\theta_i$, we select the ground RVs $\theta_{i,n}$ corresponding to the sampled indices for the plates $\operatorname{Plates}(\theta_i)$.
    \item The selected ground RVs $\theta_{i,n}$ constitute the set $\redu{\Theta}[t]$ of parameters appearing in \cref{eq:q_stoc}. The same procedure yields the observed RV subset $\redu{X}[t]$ and the data slice $\redu{\tX}[t]$.
\end{enumerate}
For instance, in the toy example from \cref{fig:PAVI_principle}, $X_2$ will be chosen iff the index $1$ is selected as part of sub-sampling $\mathcal{P}_1$ and the index $0$ is selected as part of sub-sampling $\mathcal{P}_0$.
Less formally, this is equivalent to going \textit{middle}, then \textit{left} in the full graph representing $\mathcal{M}$.
This stochastic choice is illustrated in \cref{fig:PAVI_training} at $t=1$ where $X_2$ corresponds to the node $E$.
This stochastic strategy also applies to the selected encoding scheme ---described in \cref{sec:encoding_schemes}\&\ref{sec:sharing_learning}--- as detailed in the next sections.

\subsubsection{PAVI-F details}
\label{sec:PAVI-F_details}
In \cref{sec:encoding_schemes} we refer to encodings $\tE_i = [ \tE_{i,n}]_{n=0..N}$ corresponding to RV templates $\theta_i$.
In practice, we have some amount of sharing for those encodings: instead of defining separate encodings for every RV template, we define encodings for every \textit{plate level}.
A plate level is a combination of plates with at least one parameter RV template $\theta_i$ belonging to it:
\begin{equation}
    \begin{aligned}
    \operatorname{PlateLevels} &= \{ (\mathcal{P}_k..\mathcal{P}_l) = \operatorname{Plates}(\theta_i) \}_{\theta_i \in \Theta}
    \end{aligned}
\end{equation}
For every plate level, we construct a large encoding array with the cardinalities of the full model $\mathcal{M}$:
\begin{equation}
    \begin{aligned}
    \operatorname{Encodings} &= \{ (\mathcal{P}_k..\mathcal{P}_l) \mapsto \mathbb{R}^{\operatorname{Card}(\mathcal{P}_k) \times .. \times \operatorname{Card}(\mathcal{P}_l) \times D} \}_{(\mathcal{P}_k..\mathcal{P}_l) \in \operatorname{PlateLevels}} \\
    \tE_i &= \operatorname{Encodings}(\operatorname{Plates}(\theta_i))
    \end{aligned}
\end{equation}
Where $D$ is an encoding size that we kept constant to de-clutter the notation but can vary between plate levels.
The encodings for a given ground RV $\theta_{i,n}$ then correspond to an element from the encoding array $\tE_i$. 

\subsubsection{PAVI-E details}
\label{sec:PAVI-E_details}
In the PAVI-E scheme, encodings are not free weights but the output of en encoder $f(\ \cdot \ , \eta)$ applied to the observed data $\tX$.
In this section we detail the design of this encoder.

As in the previous section, the role of the encoder will be to produce one encoding per plate level.
We start from a dependency structure for the plate levels:
\begin{equation}
    \begin{aligned}
    \forall (\mathcal{P}_a..\mathcal{P}_b) &\in \operatorname{PlateLevels} \enspace, \\
    \forall (\mathcal{P}_c..\mathcal{P}_d) &\in \operatorname{PlateLevels} \enspace, \\
    (\mathcal{P}_a..\mathcal{P}_b) \in \pi((\mathcal{P}_c..\mathcal{P}_d)) &\Leftrightarrow \substack{\exists \theta_i/\operatorname{Plates}(\theta_i) = (\mathcal{P}_a..\mathcal{P}_b) \\ \exists \theta_j/\operatorname{Plates}(\theta_j) = (\mathcal{P}_c..\mathcal{P}_d)} / \theta_j \in \pi(\theta_i)
    \end{aligned}
\end{equation}
note that this dependency structure is in the \textit{backward} direction: a plate level will be the parent of another plate level, if the former contains a RV who has a child in the latter.
We therefore obtain a plate level dependency structure that \textit{reverts} the conditional dependency structure of the graph template $\mathcal{T}$.
To avoid redundant paths in this dependency structure, we take the maximum branching of the obtained graph.

Given the plate level dependency structure, we will recursively construct the encodings, starting from the observed data:
\begin{equation}
    \begin{aligned}
    \forall x \in X:\quad \operatorname{Encodings}(\operatorname{Plates}(x)) = \rho (\textbf{x})
    \end{aligned}
\end{equation}
where $\textbf{x}$ is the observed data for the RV $x$, and $\rho$ is a simple encoder that processes every observed ground RV's value independently through an identical multi-layer perceptron.
Then, until we have exhausted all plate levels, we process existing encodings to produce new encodings:
\begin{equation}
    \begin{aligned}
    \forall (\mathcal{P}_k..\mathcal{P}_l) &\in \operatorname{PlateLevels} / \not\exists x \in X, \operatorname{Plates}(x) = (\mathcal{P}_k..\mathcal{P}_l): \\
    \operatorname{Encodings}((\mathcal{P}_k..\mathcal{P}_l)) &= g(\operatorname{Encodings}(\pi(\mathcal{P}_k..\mathcal{P}_l)))
    \end{aligned}
\end{equation}
where $g$ is the composition of attention-based deep-set networks called \textit{Set Transformers} \citep{ST, deep_sets}.
For every plate $\mathcal{P}_p$ present in the parent plate level but absent in the child plate level, $g$ will compute summary statistics \textit{across} that plate, effectively contracting the corresponding batch dimensionality in the parent encoding \citep{ADAVI}.

In the case of multiple observed RVs, we run this "backward pass" independently for each observed data ---with one encoder per observed RV.
We then concatenate the resulting encodings corresponding to the same plate level.

For more precise implementation details, we invite the reader to consult the codebase released with this supplemental material.

{\color{new}
\subsection{Analysis of bias in the stochastic training}
\label{sec:no_bias}

A key concern in our stochastic training scheme is its unbiasedness: we want our stochastic optimization to converge to the same variational posterior as if we trained over the full model directly ---without any stochasticity.
{\color{camera} In this section we first show that the PAVI-F is unbiased.
Second, we identify strategies to obtain an unbiased PAVI-E scheme, while showing how the approximations we do in practice can theoretically result in biased training.
As an important note, the negative impact of this bias on the performance PAVI-E remained limited throughout our experiments ---as seen in experiments \ref{sec:exp_scaling}, \ref{sec:exp_card_redu}, \ref{sec:sanity} and \ref{sec:sup_comparisons}.

\subsubsection{General derivation (applicable to both the PAVI-F and PAVI-E schemes)}

We first formalize the \textit{plate sampling} strategy described in \cref{sec:plate_branching}.
To every plate $\mathcal{P}$ we associate the RV $I_\mathcal{P}$ corresponding to the $\redu{\operatorname{Card}}(\mathcal{P})$-sized set of indices sampled without replacement from the $\operatorname{Card}(\mathcal{P})$ possible index values.
As an example, with a plate $\mathcal{P}_0$ with $\operatorname{Card}(\mathcal{P}_0) = 4$ and $\redu{\operatorname{Card}}(\mathcal{P}_0) = 2$, $\{ 0, 2 \}$ or $\{ 2, 3 \}$ can be 2 different samples from $I_{\mathcal{P}_0}$.
At a given optimization step $t$, we sample independently from the RVs $\{ I_{\mathcal{P}_p} \}_{p=0..P}$.
This defines the batches $\mathcal{B}_i[t]$ and the distribution $\redu{q}$ in \cref{eq:q_stoc}.

To check the unbiasedness of our stochastic training, we need to show that:
\begin{equation}
\label{eq:no_bias}
    \begin{aligned}
    \mathbb{E}_{I_{\mathcal{P}_0}} \hdots \mathbb{E}_{I_{\mathcal{P}_P}} \left[ \redu{\operatorname{ELBO}}[t] \right] &= \operatorname{ELBO}
    \end{aligned}
\end{equation}
Where:
\begin{equation}
    \begin{aligned}
    \operatorname{ELBO} &= \mathbb{E}_{\Theta \sim q}\left[ \log p(X, \Theta) - \log q(\Theta) \right]
    \end{aligned}
\end{equation}
And $\redu{\operatorname{ELBO}}[t]$ is defined in \cref{eq:ELBO_stoc}.
In that expression, $q$ and $p$ have symmetrical roles.
As the ELBO amounts to the difference between the logarithms of distributions $p$ and $q$, we can prove the equality in \cref{eq:no_bias} if we prove that the expectation of each reduced distribution is equal to the corresponding full distribution.
To prove the equality in \cref{eq:no_bias}, a sufficient condition is therefore to prove that:
\begin{equation}
\label{eq:no_bias_q}
    \mathbb{E}_{I_{\mathcal{P}_0}} \hdots \mathbb{E}_{I_{\mathcal{P}_P}} \left[ \log \redu{q}(\redu{\Theta}[t]) \right]
        = \mathbb{E}_{I_{\mathcal{P}}} \left[ \log \redu{q}(\redu{\Theta}[t]) \right]
        = \log q(\Theta)
\end{equation}
where to de-clutter the notations we denote the expectation over the collection of RVs $\{ I_{\mathcal{P}_p} \}_{p=0..P}$ as $\mathbb{E}_{I_\mathcal{P}}$.

Consider a given ground RV $\theta_{i, n}$ corresponding to the RV template $\theta_i$ and to the plates $\operatorname{Plates}(\theta_i)$.
At a given stochastic step $t$, $\theta_{i, n}$ will be chosen if and only if its corresponding \textit{branching} is chosen.
Recall that when sampling equiprobably without replacement a set of $k$ elements from a population of $n$ elements, a given element will be present in the set with probability $k / n$.
We can apply this reasoning to the choice of \textit{branching} corresponding to a given ground RV.
For instance, in \cref{fig:PAVI_principle}, $X_2$ will be chosen iff the index $1$ is selected as part of sub-sampling $\mathcal{P}_1$ and the index $0$ is selected as part of sub-sampling $\mathcal{P}_0$.
As $\redu{\operatorname{Card}}(\mathcal{P}_1)=2$ indices are chosen inside the plate $\mathcal{P}_1$ of full cardinality 3, and $\redu{\operatorname{Card}}(\mathcal{P}_0)=1$ indices are chosen inside the plate $\mathcal{P}_0$ of full cardinality 2, $X_2$ is therefore chosen with probability $2/3 \times 1/2$.
More formally, for a ground RV $\theta_{i,n}$ we have:
\begin{equation}
    \begin{aligned}
    \forall n = 0 .. N_i\ : \ \mathbb{P}(\theta_{i, n} \in \mathcal{B}_i[t]) &= \prod_{\mathcal{P} \in \operatorname{Plates}(\theta_i)} \frac{\redu{\operatorname{Card}}(\mathcal{P})}{\operatorname{Card}(\mathcal{P})} \\
            &= \frac{\redu{N_i}}{N_i}
    \end{aligned}
\end{equation}
Applying this reasoning to every RV template $\theta_i$, we have that:
\begin{equation}
    \begin{aligned}
    \mathbb{E}_{I_{\mathcal{P}}} \left[ \log \redu{q}(\redu{\Theta}[t]) \right]
            &= \sum_{i=1}^I \frac{N_i}{\redu{N_i}} \mathbb{E}_{I_{\mathcal{P}}} \left[ \sum_{\mathclap{\qquad n \in \mathcal{B}_{i}[t]}} \log q_{i,n}(\theta_{i,n} | \pi(\theta_{i,n})) \right] \\
            &= \sum_{i=1}^I \frac{N_i}{\redu{N_i}} \mathbb{E}_{I_{\mathcal{P}}} \left[ \sum_{n=0}^{N_i} \mathbbm{1}_{n \in \mathcal{B}_{i}[t]} \log q_{i,n}(\theta_{i,n} | \pi(\theta_{i,n})) \right] \\
            &= \sum_{i=1}^I \frac{N_i}{\redu{N_i}} \sum_{n=0}^{N_i} \mathbb{E}_{I_{\mathcal{P}}} \left[\mathbbm{1}_{n \in \mathcal{B}_{i}[t]} \log q_{i,n}(\theta_{i,n} | \pi(\theta_{i,n}); \phi_i, \tE_{i,n}) \right]
    \end{aligned}
\end{equation}
{\color{camera}
where we exploited the fact that the expectation of the sum of RVs is the sum of the expectations, even in the case of dependent RVs.
The term $\mathbbm{1}_{n \in \mathcal{B}_{i}[t]} \times \log q_{i,n}(\theta_{i,n} | \pi(\theta_{i,n}); \phi_i, \tE_{i,n})$ is the product of 2 RVs ---related to the stochastic choice of plate indices:
\begin{itemize}
    \item the RV $\mathbbm{1}_{n \in \mathcal{B}_{i}[t]}$ is an indicator that $\theta_{i,n}$'s \textit{branching} has been chosen via the stochastic sampling of plate indices.
    By construction, this RV depends only on the indices of the plates $\mathcal{P} \in \operatorname{Plates}(\theta_i)$.
    \item the RV $\log q_{i,n}(\theta_{i,n} | \pi(\theta_{i,n}); \phi_i, \tE_{i,n})$ depends on $\tE_{i,n}$, whose construction depends on the encoding scheme:
    \begin{itemize}
        \item In the PAVI-F scheme, $\tE_{i,n}$ is a constant.
        \item In the PAVI-E scheme, $\tE_{i,n}$ results of the application of an encoder to the observed data of a subset of $\theta_{i,n}$'s descendants.
        By construction, this subset will only depend on the indices of plates containing $\theta_i$'s descendants, but not containing $\theta_i$.
        The value of $\tE_{i,n}$ therefore only depends on the indices of plates $\mathcal{P}\notin \operatorname{Plates}(\theta_i)$
    \end{itemize}
\end{itemize}
\rebutal{As an example of this reasoning, consider the model $\mathcal{M}$ illustrated in \cref{fig:PAVI_training}.
We can evaluate both terms for the ground RV $\theta_{1, 2}$ in the PAVI-E scheme:
\begin{itemize}
    \item $\mathbbm{1}_{2 \in \mathcal{B}_{1}[t]}$ depends on whether the index $2$ is chosen as part of sub-sampling the plate $\mathcal{P}_1$, and therefore only depends on the RV $I_{\mathcal{P}_1}$. In this case, the associated probability is $2/3$;
    \item to evaluate $\log q_{1,2}(\theta_{1,2} | \theta_{2,0}; \phi_1, \tE_{1,2})$, the value of $\tE_{1, 2}$ will result from the application of the encoder $f$ over the value of either $X_4$ or $X_5$.
    This choice depends on whether the index $0$ or $1$ is chosen as part of sub-sampling the plate $\mathcal{P}_0$.
    Therefore, the value of the term $\log q_{1,2}$ only depends on the RV $I_{\mathcal{P}_0}$.
\end{itemize}}

As a result, in both PAVI-F and PAVI-E, the terms $\mathbbm{1}_{n \in \mathcal{B}_{i}[t]}$ and $\log q_{i,n}(\theta_{i,n} | \pi(\theta_{i,n}); \phi_i, \tE_{i,n})$ depend on the sampled indices of disjoint sets of plates, and are therefore independent.
This means that the expectation of their product can be rewritten as the product of their expectations:
\begin{equation}
\label{eq:E_q_redu}
    \begin{aligned}
     \mathbb{E}_{I_{\mathcal{P}}}  \left[ \log \redu{q}(\redu{\Theta}[t]) \right]
        &= \sum_{i=1}^I \frac{N_i}{\redu{N_i}} \sum_{n=0}^{N_i} \mathbb{E}_{I_{\mathcal{P}}} \left[\mathbbm{1}_{n \in \mathcal{B}_{i}[t]}\right] \mathbb{E}_{I_{\mathcal{P}}} \left[ \log q_{i,n}(\theta_{i,n} | \pi(\theta_{i,n})) \right] \\
        &= \sum_{i=1}^I \frac{N_i}{\redu{N_i}} \sum_{n=0}^{N_i} \frac{\redu{N_i}}{N_i} \mathbb{E}_{I_{\mathcal{P}}} \left[\log q_{i,n}(\theta_{i,n} | \pi(\theta_{i,n}))\right] \\
        &= \sum_{i=1}^I \sum_{n=0}^{N_i} \mathbb{E}_{I_{\mathcal{P}}} \left[\log q_{i,n}(\theta_{i,n} | \pi(\theta_{i,n}); \tE_{i,n})\right]
    \end{aligned}
\end{equation}
This equality can be further simplified in the PAVI-F case ---proving its unbiasedness--- but not in the PAVI-E case, as detailed in the sections below.

\subsubsection{Unbiasedness of the PAVI-F scheme}
In the PAVI-F scheme, detailed in \cref{sec:encoding_schemes}, the encodings $\tE_{i,n}$ are constants with respect to the branching choice, therefore we have:
\begin{equation}
\label{eq:unbiased_PAVI-F}
\begin{aligned}
\mathbb{E}_{I_{\mathcal{P}}} \left[ \log \redu{q}(\redu{\Theta}[t]) \right]
    &= \sum_{i=1}^I \sum_{n=0}^{N_i} \mathbb{E}_{I_{\mathcal{P}}} \left[\log q_{i,n}(\theta_{i,n} | \pi(\theta_{i,n}); \tE_{i,n})\right] \\
    &= \sum_{i=1}^I \sum_{n=0}^{N_i} \log q_{i,n}(\theta_{i,n} | \pi(\theta_{i,n}); \tE_{i,n}) \\
    &= \log q(\Theta)
\end{aligned}
\end{equation}
which proves \cref{eq:no_bias_q} and \cref{eq:no_bias}.
\rebutal{In the above example of $\theta_{1,2}$ in $\mathcal{M}$, in the PAVI-F scheme the expression  $ \mathbb{E}_{I_{\mathcal{P}}} \left[\log q_{i,n}(\theta_{i,n} | \pi(\theta_{i,n}); \phi_i, \tE_{i,n})\right]$ can be evaluated into $\log q_{1,2}(\theta_{1,2} | \theta_{2,0}; \phi_1, \tE_{1,2})$.}
This demonstrates that the PAVI-F scheme is unbiased: training over stochastically chosen sub-graphs for $\redu{q}$ is in expectation equal to training over the full graph of $q$.

\subsubsection{Approximations in the PAVI-E scheme}
In the PAVI-E scheme, detailed in \cref{sec:sharing_learning}, the encodings $\tE_{i,n}$ are computed from the observed data $X$.
Specifically, considering the ground RV $\theta_{i,n}$, we have $\tE_{i,n} = f(\redu{\tX}_{i,n}[t])$ where $\redu{\tX}_{i,n}[t]$ corresponds to the observed data of a subset of $\theta_{i,n}$'s descendants.
Depending on the chosen branching \textit{downstream} of $\theta_{i,n}$, the value of $\tE_{i,n}$ can therefore vary.
This means we cannot further simplify \cref{eq:E_q_redu}: the terms $\log q_{i,n}(\theta_{i,n} | \pi(\theta_{i,n}); \tE_{i,n})$ are not constants with respect to the RVs $I_{\mathcal{P}}$.
\rebutal{In the above example of $\theta_{1,2}$ in $\mathcal{M}$, in the PAVI-E scheme the expression  $ \mathbb{E}_{I_{\mathcal{P}}} \left[\log q_{i,n}(\theta_{i,n} | \pi(\theta_{i,n}); \phi_i, \tE_{i,n})\right]$ can be evaluated into: 
\begin{equation*}
    \frac{1}{2}(\log q_{1,2}(\theta_{1,2} | \theta_{2,0}; \phi_1, f(\tX_4)) + \log q_{1,2}(\theta_{1,2} | \theta_{2,0}; \phi_1, f(\tX_5)))
\end{equation*}}

\textbf{How could the PAVI-E scheme be made unbiased?}
Specifically, by making the value of $\tE_{i,n}$ independent of the choice of downstream branching.
A possibility would be to parameterize $\tE_{i,n}$ as an average ---an expectation--- over all the possible sub-branchings downstream of $\theta_{i,n}$.
Yet, in practical cases, the cardinalities of the reduced model are much inferior to the ones of the full model: $\redu{\operatorname{Card}}(\mathcal{P}) \ll \operatorname{Card}(\mathcal{P})$.
This means that numerous $\redu{\operatorname{Card}}(\mathcal{P})$-sized subsets can be chosen inside the $\operatorname{Card}(\mathcal{P})$ possible descendants.
In order to average over all those subset choices to compute $\tE_{i,n}$, numerous encoding calculations would be required at each stochastic training step.
For large-scale cases, we deemed this possibility impractical.
Other possibilities could exist, all revolving around the problem of aggregating collections of stochastic estimators into one general estimator ---in an unbiased and efficient manner.
To our knowledge, this is a complex and still open research question, whose advancement could much benefit our applications.

\textbf{Practical approximation for the PAVI-E scheme}
In practice, we compute the encoding $\tE_{i,n}$ based on the single downstream branching corresponding to the sampling of the RVs $I_\mathcal{P}$.
Compared to the previous paragraph, this amounts to estimating the expectation of $\tE_{i,n}$ ---over all downstream branchings--- using a single one of those branchings.
Note that, even if this encoding estimate was unbiased, $\log q_{i,n}$ would remain a highly non-linear function of $\tE_{i,n}$.
As a consequence, we need to rely on the approximation:
\begin{equation}
    \label{eq:approx_PAVI-E}
    \mathbb{E}_{I_{\mathcal{P}}} \left[\log q_{i,n}(\theta_{i,n} | \pi(\theta_{i,n}); \phi_i, f(\redu{\tX_{i,n}}[t]))\right] \simeq \log q_{i,n}(\theta_{i,n} | \pi(\theta_{i,n}); \phi_i, f(\tX_{i,n}))
\end{equation}
which can theoretically introduce some bias in our gradients.
The approximation \cref{eq:approx_PAVI-E} can be interpreted as follow: "the expectation of the density of $\theta_{i,n}$ when collecting summary statistics over a stochastic subset of $\theta_{i,n}$'s descendants is approximately equal to the density of $\theta_{i,n}$ when collecting summary statistics over the entirety of $\theta_{i,n}$'s descendants".
Another interpretation is that the distribution associated with the summary of the full data can be approximated by annealing the distributions associated with summaries of subsets of this data.
In practice, this approximation did not yield significantly worse performance for the PAVI-E scheme over the generative models we tested.
At the same time, computing the encodings over a single branching allows the computation of all the $\tE_{i,n}$ encodings in a single lightweight pass over the data $\redu{\tX}[t]$.
This simple solution thus provided a substantial increase in training speed with seldom noticeable bias.
Yet, we do not bar the existence of pathological generative HBMs where this approximation would become coarse.
Experimenters should bear in mind this possibility when using the PAVI-E scheme.
In practice, using the PAVI-F scheme as a sanity check over synthetic, toy-dimension implementations of the considered generative models is a good way to validate the PAVI-E scheme ---before moving on to the real problem instantiating the same generative model with a larger dimensionality.
}

}
\subsection{PAVI algorithms}

More technical details can be found in the codebase provided with this supplemental material.

\subsubsection{Architecture build}
\begin{algorithm}[H]
\caption{PAVI architecture build}
\label{alg:arch}
\KwIn{Graph template $\mathcal{T}$, plate cardinalities $\{ (\operatorname{Card}(\mathcal{P}_p), \redu{\operatorname{Card}}(\mathcal{P}_p)) \}_{p=0..P}$, encoding scheme}
\KwOut{$q$ distribution}
\For{$i=1..I$}{
    Construct conditional flow $\mathcal{F}_i$\;
    Define conditional posterior distributions $q_{i,n}$ as the push-forward of the prior via $\mathcal{F}_i$, following \cref{eq:q_full}\;
}
Combine the $q_{i,n}$ distributions following the cascading flows scheme, as in \cref{sec:variational_family} \citep{CF} \;
\uIf{PAVI-F encoding scheme}{
    Construct encoding arrays $\{  \tE_i = \left[ \tE_{i,n} \right]_{n=0..N_i} \}_{i=1..I}$ as in \cref{sec:PAVI-F_details} \;
}
\ElseIf{PAVI-E encoding scheme}{
    Construct encoder $f$ as in \cref{sec:PAVI-E_details} \;
}
\end{algorithm}

\subsubsection{Stochastic training}
\begin{algorithm}[H]
\caption{PAVI stochastic training}
\label{alg:train}
\KwIn{Untrained architecture $q$, observed data $\tX$, encoding scheme, number of steps $T$}
\KwOut{trained architecture $q$}
\For{$t=0..T$}{
    Sample plate indices to define the batches $\mathcal{B}_i[t]$, the latent $\redu{\Theta}[t]$ and the observed $\redu{X}[t]$ and $\redu{\tX}[t]$, following \cref{sec:plate_branching} \;
    Define reduced distribution $\redu{p}$ \;
    \uIf{PAVI-F encoding scheme}{
        Collect encodings $\tE_{i,n}$ by slicing from the arrays $\tE_i$ the elements corresponding to the batches $\mathcal{B}_i[t]$ \;
    }
    \ElseIf{PAVI-E encoding scheme}{
        Compute encodings as $\tE = f(\redu{\tX}[t]) $\;
    }
    Feed obtained encodings into $\redu{q}$ \;
    Compute reduced ELBO as in \cref{eq:ELBO_stoc}, back-propagate its gradient \;
    Update conditional flow weights $\{ \phi_i \}_{i=1..I}$\;
    \uIf{PAVI-F encoding scheme}{
        Update encodings $\{ \tE_{i,n} \}_{i=1..I, n \in \mathcal{B}_{i, t}}$\;
    }
    \ElseIf{PAVI-E encoding scheme}{
        Update encoder weights $\eta$\;
    }
}
\end{algorithm}

\subsubsection{Inference}
\begin{algorithm}[H]
\caption{PAVI inference}
\label{alg:infer}
\KwIn{trained architecture $q$, observed data $\tX$, encoding scheme}
\KwOut{approximate posterior distribution}
\uIf{PAVI-F encoding scheme}{
    Collect full encoding arrays $\tE_i$ \;
}
\ElseIf{PAVI-E encoding scheme}{
    Compute encodings as $\tE = f(\tX)$ using set size generalization \;
}
Feed obtained encodings into $q$ \;
\end{algorithm}

\subsection{Inference gaps}
\label{sec:inference_gap}
In terms of inference quality, the impact of our architecture can be formalized following the \textit{gaps} terminology \citep{amortization_gap}.
Consider a joint distribution $p(\Theta, X)$, and a value $\tX$ for the RV template $X$.
We pick a variational family $\mathcal{Q}$, and in this family look for the parametric distribution $q(\Theta; \phi)$ that best approximates $p(\Theta | X=\tX)$.
Specifically, we want to minimize the Kulback-Leibler divergence \citep{blei_variational_2017} between our variational posterior and the true posterior, that \citet{amortization_gap} refer to as the \textit{gap} $\mathcal{G}$:
\begin{equation}
\begin{aligned}
    \mathcal{G} &= \operatorname{KL}(q(\Theta;\phi) || p(\Theta | X)) \\
            &= \log p(X) - \operatorname{ELBO}(q; \phi)
\end{aligned}
\end{equation}
We denote $q^{*}(\Theta; \phi^*)$ the optimal distribution inside $\mathcal{Q}$ that minimizes the KL divergence with the true posterior:
\begin{equation}
    \begin{aligned}
    \mathcal{G}_{\text{approx}}(\mathcal{Q}; \phi^*) &= \log p(X) - \operatorname{ELBO}(q^*; \phi^*) \\
    &\geq 0 \\
    \mathcal{G}_{\text{vanilla VI}} &= \mathcal{G}_{\text{approx}}
    \end{aligned}
\end{equation}
The \textit{approximation gap} $\mathcal{G}_{\text{approx}}$ depends on the expressivity of the variational family $\mathcal{Q}$, specifically whether $\mathcal{Q}$ contains distributions arbitrarily close to the posterior ---in the KL sense.

\rebutal{\textbf{Note:} $\mathcal{G}_{\text{approx}}$ is a property of the variational family $\mathcal{Q}$.
$\mathcal{G}_{\text{approx}}$ is an asymptotic bound for the KL divergence between any distribution $q\in \mathcal{Q}$ and the true posterior.
This gap is therefore a form of bias but is not to be mistaken with the stochasticity-induced bias studied in \cref{sec:no_bias}.
The bias in \cref{sec:no_bias} relates to whether $q^*$ can be found by training stochastically over $\mathcal{Q}$, whereas $\mathcal{G}_{\text{approx}}$ relates to the the bias between $q^*$ and the true posterior.}

\citet{amortization_gap} demonstrate that, in the case of sample amortized inference, when the weights $\phi$ no longer are free but the output of an encoder $f \in \mathcal{F}$, inference cannot be better than in the non-sample-amortized case, and a positive \textit{amortization gap} is introduced:
\begin{equation}
\label{eq:sa}
    \begin{aligned}
    \mathcal{G}_{\text{sa}}(\mathcal{Q}, \mathcal{F}; \eta^*) &= \mathcal{G}_{\text{approx}}(\mathcal{Q}; f(\tX, \eta^*)) - \mathcal{G}_{\text{approx}}(\mathcal{Q}; \phi^*) \\
    &\geq 0 \\
    \mathcal{G}_{\text{sample amortized VI}} &= \mathcal{G}_{\text{approx}} + \mathcal{G}_{\text{sa}}
    \end{aligned}
\end{equation}
Where we denote as $\eta^*$ the optimal weights for the encoder $f$ inside the function family $\mathcal{F}$.
The gap terminology can be interpreted as follow: "theoretically, sample amortization cannot be beneficial in terms of KL divergence for the inference over a given sample $\tX$."

Using the same gap terminology, we can define gaps implied by our PAVI architecture.
Instead of picking the distribution $q$ inside the family $\mathcal{Q}$, consider picking $q$ from the \textit{plate-amortized} family $\mathcal{Q}_{\text{pa}}$ corresponding to $\mathcal{Q}$. Distributions in $\mathcal{Q}_{\text{pa}}$ are distributions from $\mathcal{Q}$ with the additional constraints that some weights have to be equal. Consequently, $\mathcal{Q}_{\text{pa}}$ is a subset of $\mathcal{Q}$:
\begin{equation}
    \mathcal{Q}_{\text{pa}} \subset \mathcal{Q}
\end{equation}
As such, looking for the optimal distribution inside $\mathcal{Q}_{\text{pa}}$ instead of inside $\mathcal{Q}$ cannot result in better performance, leading to a \textit{plate amortization gap}:
\begin{equation}
\label{eq:pa}
    \begin{aligned}
    \mathcal{G}_{\text{pa}}(\mathcal{Q}, \mathcal{Q}_{\text{pa}}; \psi^*, \phi^*) &=  \mathcal{G}_{\text{approx}}(\mathcal{Q}_{\text{pa}}; \psi^*) - \mathcal{G}_{\text{approx}}(\mathcal{Q}; \phi^*) \\
    &\geq 0 \\
    \mathcal{G}_{\text{PAVI-F}} &= \mathcal{G}_{\text{approx}} + \mathcal{G}_{\text{pa}}
    \end{aligned}
\end{equation}
Where we denote as $\psi^*$ the optimal weights for a variational distribution $q$ inside $\mathcal{Q}_{\text{pa}}$ ---in the KL sense. The equation \ref{eq:pa} is valid for the PAVI-F scheme ---see \cref{sec:encoding_schemes}. We can interpret it as follow: "theoretically, plate amortization cannot be beneficial in terms of KL divergence for the inference over a given sample $\tX$".

Now consider that encodings are no longer free parameters but the output of an encoder $f$. Similar to the case presented in \cref{eq:sa}, using an encoder cannot result in better performance, leading to an \textit{encoder gap}:
\begin{equation}
\label{eq:pa-e}
    \begin{aligned}
    \mathcal{G}_{\text{encoder}}(\mathcal{Q}_{\text{pa}}, \mathcal{F}; \psi^*, \eta^*) &=  \mathcal{G}_{\text{approx}}(\mathcal{Q}_{\text{pa}}; f(\tX, \eta^*)) - \mathcal{G}_{\text{approx}}(\mathcal{Q}_{\text{pa}}; \psi^*) \\
    &\geq 0 \\
    \mathcal{G}_{\text{PAVI-E}} &= \mathcal{G}_{\text{approx}} + \mathcal{G}_{\text{pa}} + \mathcal{G}_{\text{encoder}}
    \end{aligned}
\end{equation}
The equation \cref{eq:pa-e} is valid for the PAVI-E scheme ---see \cref{sec:encoding_schemes}.

The most complex case is the PAVI-E(sa) scheme, where we combine both plate and sample amortization.
Our argument cannot account for the resulting $\mathcal{G}_{\text{PAVI-E(sa)}}$ gap: both the PAVI-E and PAVI-E(sa) schemes rely upon the same encoder $f$.
In the PAVI-E scheme, $f$ is overfitting over a dataset composed of the slices of a given data sample $\tX$.
In the PAVI-E(sa) scheme, the encoder is trained  over the whole distribution of the samples of the reduced model $\redu{\mathcal{M}}$.
Intuitively, it is likely that the performance of PAVI-E(sa) will always be dominated by the performance of PAVI-E, but ---as far as we understand it--- the gap terminology cannot account for this discrepancy.

Comparing previous equations, we therefore have:
\begin{equation}
    \mathcal{G}_{\text{vanilla VI}} \leq \mathcal{G}_{\text{PAVI-F}} \leq \mathcal{G}_{\text{PAVI-E}}
\end{equation}
Note that those are \textit{theoretical} results, that do not necessarily pertain to optimization in practice.
In particular, in \cref{sec:exp_convergence_speed}\&\ref{sec:exp_scaling}, this theoretical performance loss is not observed empirically over the studied examples.
On the contrary, in practice, our results can actually be better than non-amortized baselines, as is the case for the PAVI-F scheme in \cref{fig:scaling} or experiments \ref{sec:sup_comparisons}.
We interpret this as a result of a simplified optimization problem due to plate amortization ---with fewer parameters to optimize for, and mini-batching effects across different ground RVs.
A better framework to explain those discrepancies could be the one from \citet{bottou_tradeoffs_2007}: performance in practice is not only the reflection of an \textit{approximation error} but also of an \textit{optimization error}.
A less expressive architecture ---using plate amortization--- may in practice yield better performance.
Furthermore, for the experimenter, the theoretical gaps $\mathcal{G}_{\text{pa}}, \mathcal{G}_{\text{encoder}}$ are likely to be well "compensated for" by the lighter parameterization and faster convergence entitled by plate amortization.
\section{Supplemental results}

In this section, we present various supplemental experiments that further compare the PAVI-E and -F between themselves and against baselines.
Overall, we found the PAVI-F scheme faster to train and to yield better inference quality than the PAVI-E scheme.
When its parameterization is affordable, PAVI-F should be preferred.
PAVI-E nevertheless opens up promising research directions, with the potential for parameterization-constant, time-constant sample-amortized inference as the problem cardinality augments.
Though degraded compared to PAVI-F's, PAVI-E's performance is still on par or beats baselines in a variety of inference tasks ---see experiments \ref{sec:exp_scaling}\&\ref{sec:sup_comparisons}.

\tmlrrebutal{
\subsection{Tabular results for experiment \ref{sec:exp_scaling}}
For convenience, we reproduce the results of our comparative experiment from \cref{fig:scaling} in \cref{tab:scaling}.
}

\begin{table}
    \centering
    \begin{tabular}{rrlll}
         & Architecture & ELBO ($10^1$) & Parameterization & Optimization time (s) \\
        \\ \hline \\
        $\operatorname{Card}(\mathcal{P}_1) = 2$ & ADAVI (sa) & 1.88 ($\pm$ 0.21) & 12,280 & 503 ($\pm$ 50) \\
         & CF (sa) & 1.48 ($\pm$ 0.05) & 2,275 & 228 ($\pm$ 23) \\
         & CF & 1.98 ($\pm$ 0.02) & 355 & 63 ($\pm$ 6) \\
         & UIVI & 2.11 ($\pm$ 0.05) & 2,764 & 220 ($\pm$ 22) \\
         & PAVI-F & 2.14 ($\pm$ 0.02) & 4,058 & 65 ($\pm$ 6) \\
         & PAVI-E (sa) & 2.03 ($\pm$ 0.03) & 27,394 & 140 ($\pm$ 14) \\
         & PAVI-E & 1.74 ($\pm$ 0.18) & 27,394 & 37 ($\pm$ 4) \\
         & \textit{Asymptotic ELBO} & 2.20 & & \\
         \\ \hline \\
         $\operatorname{Card}(\mathcal{P}_1) = 20$ & ADAVI (sa) & 21.6 ($\pm$ 1.36) & 12,280 & 800 ($\pm$ 80) \\
         & CF (sa) & 17.3 ($\pm$ 0.60) & 21,751 & 1,612 ($\pm$ 161) \\
         & CF & 22.6 ($\pm$ 0.01) & 2,551 & 404 ($\pm$ 40) \\
         & UIVI & 23.9 ($\pm$ 0.02) & 14,676 & 230 ($\pm$ 23) \\
         & PAVI-F & 23.8 ($\pm$ 0.02) & 4,202 & 72 ($\pm$ 7) \\
         & PAVI-E (sa) & 22.3 ($\pm$ 0.30) & 27,394 & 145 ($\pm$ 14) \\
         & PAVI-E & 18.4 ($\pm$ 2.63) & 27,394 & 50 ($\pm$ 5) \\
         & \textit{Asymptotic ELBO} & 24.1   & & \\
         \\ \hline \\
         $\operatorname{Card}(\mathcal{P}_1) = 200$ & ADAVI (sa) & 221.7 ($\pm$ 10.93) & 12,280 & 1,600 ($\pm$ 160) \\
         & CF (sa) & 175.7 ($\pm$ 0.52) & 216,511 & 12,000 ($\pm$ 1,200) \\
         & CF & 228.6 ($\pm$ 0.52) & 24,511 & 2,600 ($\pm$ 260) \\
         & UIVI & 241.2 ($\pm$ 0.14) & 139,300 & 240 ($\pm$ 24) \\
         & PAVI-F & 237.7 ($\pm$ 0.95) & 5,642 & 81 ($\pm$ 8) \\
         & PAVI-E (sa) & 227.0 ($\pm$ 4.76) & 27,394 & 150 ($\pm$ 15) \\
         & PAVI-E & 224.2 ($\pm$ 15.34) & 27,394 & 85 ($\pm$ 9) \\
         & \textit{Asymptotic ELBO} & 243.4 & & \\
    \end{tabular}
    \caption{
    \tmlrrebutal{\textbf{Scaling with plate cardinalities} Reproduction of the results from \cref{fig:scaling}}}
    \label{tab:scaling}
\end{table}

\subsection{GRE results sanity check}
\label{sec:sanity}
As exposed in the introduction of \cref{sec:exps}, in this work we focused on the usage of the ELBO as an inference performance metric \citep{blei_variational_2017}:
\begin{equation}
    \begin{aligned}
    \operatorname{ELBO}(q) &= \log p(X) - \operatorname{KL}(q(\Theta) || p(\Theta | X))
    \end{aligned}
\end{equation}
Given that the likelihood term $\log p(X)$ does not depend on the variational family $q$, differences in ELBOs directly transfer to differences in KL divergence, and provide a straightforward metric to compare different variational posteriors.
Nonetheless, the ELBO doesn't provide an absolute metric of quality.
As a sanity check, we want to assert the quality of the results presented in \cref{sec:exp_scaling} ---that are transferable to \cref{sec:exp_convergence_speed}\&\ref{sec:exp_encoding_size}, based on the same model.
In \cref{fig:sanity} we plot the posterior samples of various methods against approximate closed-form ground truths, using the $\operatorname{Card}(\mathcal{P}_1) = 20$ case.
All the method's results are aligned with the closed-form ground truth, with differences in ELBO translating meaningful qualitative differences in terms of inference quality.

\begin{figure}
    \centering
    \includegraphics[height=0.73\textheight]{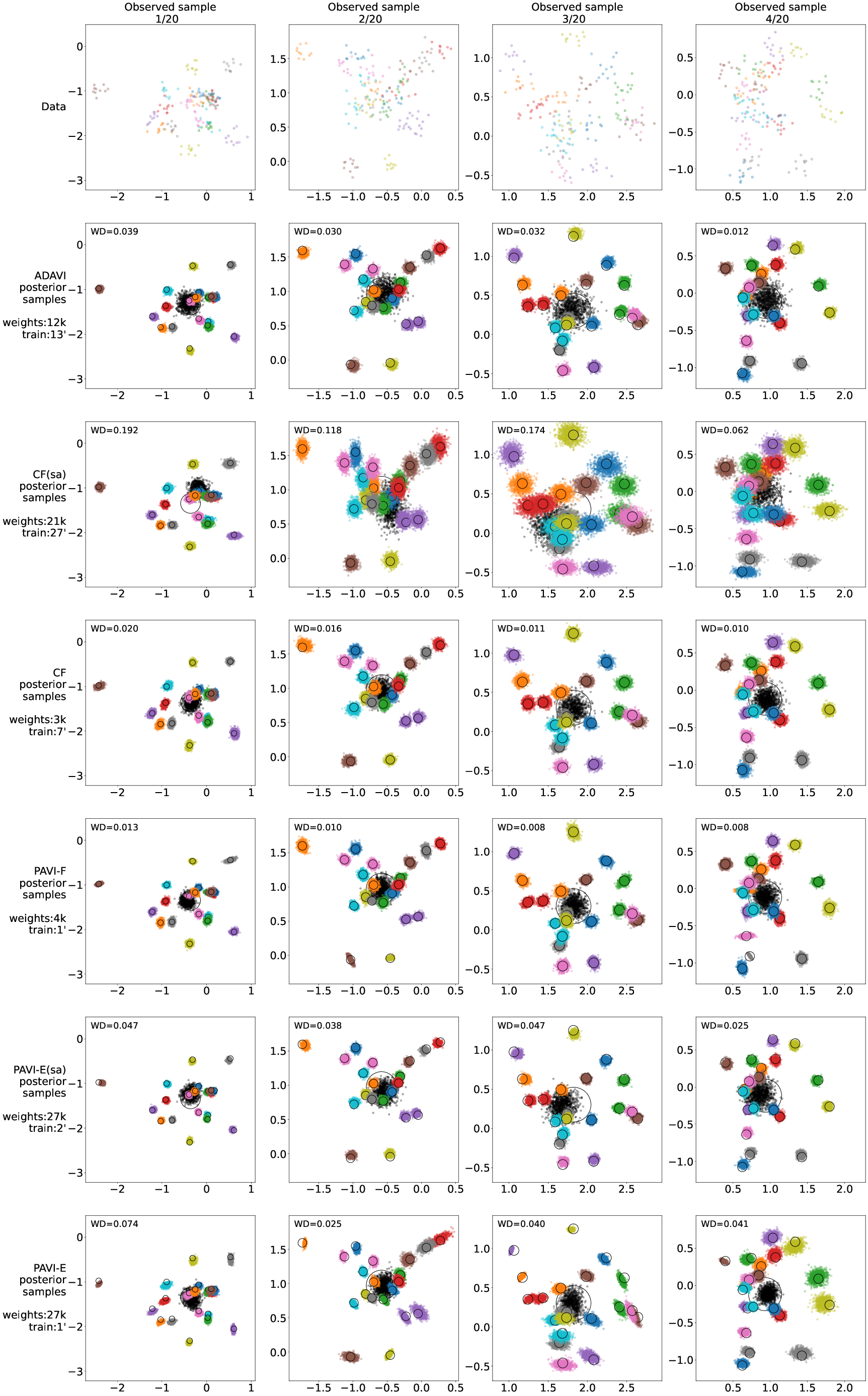}
    \caption{\textbf{GRE Sanity check} Inference methods present qualitatively correct results, making ELBO comparisons relevant in our experiments. \textit{On the topmost line}, we represent 4 different $\tX$ samples for the GRE model described in \cref{eq:GRE} with $\operatorname{Card}(\mathcal{P}_1) = 20$.
    Each set of colored points represents the $\tX_{n_1, \bullet}$ points belonging to one of the 20 groups.
    \textit{Bottom lines} represent the posterior samples for the methods used in \cref{sec:exp_scaling}.
    Colored points are sampled from the posterior of the groups means $\theta_1$, whereas black points are samples from the population mean $\theta_2$.
    We represent as black circles a closed-form ground truth, centered on the correct posterior mean, and with a radius equal to 2 times the closed-form posterior's standard deviation.
    \textbf{Correct posterior samples should be centered on the same point as the corresponding black circle, and 95\% of the points should fall within the black circle}.
    \tmlrrebutal{Associated with each posterior sample is a Wasserstein Distance (WD) computed with a sample from an approximate closed-form posterior~\citep{flamary2021pot}.}
    PAVI is represented on the 3 last lines.
    Some minor bias can be observed for the PAVI-E scheme, but this approximation error is marginal compared to the optimization error that can be observed for unbiased methods, such as CF(sa) \citep{bottou_tradeoffs_2007}.
    We can observe a superior quality for the PAVI-F scheme, rivaling ADAVI and CF's performance with orders of magnitude fewer parameters and training time, as visible in \cref{fig:scaling}.}
    \label{fig:sanity}
\end{figure}

{\color{new}
\subsection{Effect of the reduced model cardinalities on the training efficiency}
\label{sec:exp_card_redu}

\begin{figure}
    \begin{subfigure}[t]{0.48\textwidth}
        \centering
        \includegraphics[width=\textwidth]{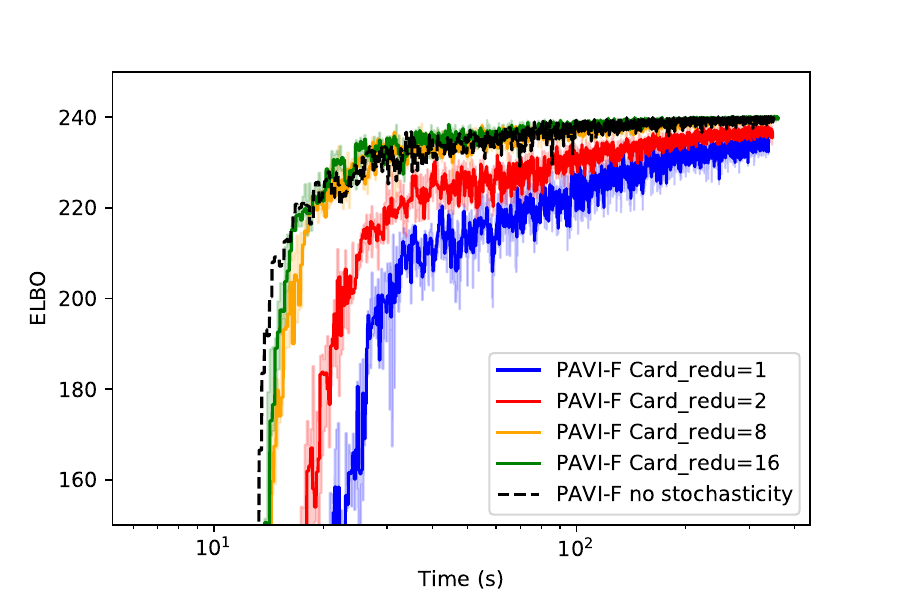}
    \end{subfigure}
    \hfill
    \begin{subfigure}[t]{0.48\textwidth}
        \centering
        \includegraphics[width=\textwidth]{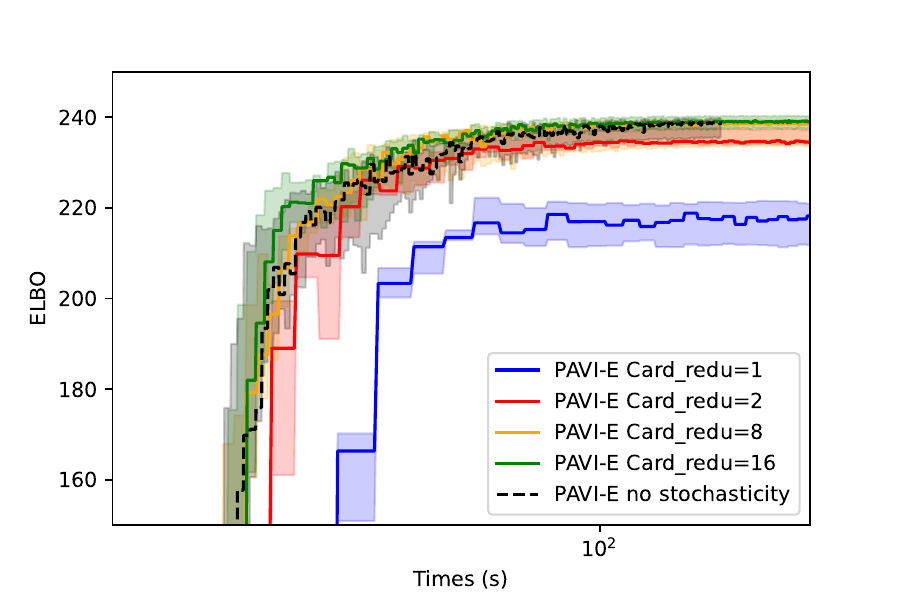}
    \end{subfigure}
    \caption{{\color{new}
    \textbf{Both panels: Effect of $\redu{\operatorname{Card}}(\mathcal{P}_1)$ on the training efficiency}
    Experiment performed on the GRE model (see \cref{eq:GRE}).
    We increase the cardinality of the reduced model $\redu{\operatorname{Card}}(\mathcal{P}_1)$ from 1 to 20 while keeping $\operatorname{Card}(\mathcal{P}_1) = 20$ fixed.
    At $\redu{\operatorname{Card}}(\mathcal{P}_1)=20$, there is no stochasticity in the training, meaning we train directly on $\mathcal{M}$.
    This experiment is interesting to evaluate the bias introduced in the stochastic training, as detailed in \cref{sec:no_bias}.
    \textbf{Left panel: PAVI-F scheme}
    As $\redu{\operatorname{Card}}(\mathcal{P}_1)$ augments, the training gets faster and less noisy, likely due to less stochasticity in the gradient estimates and more encodings vectors $\tE_{i,n}$ being trained at once.
    This increase in speed quickly caps, and the speed is approximately the same between $\redu{\operatorname{Card}}(\mathcal{P}_1)=8$ and $\redu{\operatorname{Card}}(\mathcal{P}_1)=16$.
    This experiment also illustrates the unbiasedness of the stochastic training: at $\redu{\operatorname{Card}}(\mathcal{P}_1)=20$ there is no stochasticity in the training, and the asymptotic performance is the same as for the stochastic training (see \cref{sec:no_bias}).
    \textbf{Right panel: PAVI-E scheme}
    The case of $\redu{\operatorname{Card}}(\mathcal{P}_1)=1$ is pathological: the encoder "learns" to collect summary statistics across a set of 1 element, and ---not surprisingly--- the learned function doesn't generalize well on sets of size 20.
    In all of the other cases, the ELBO converges approximately to the same values as with $\redu{\operatorname{Card}}(\mathcal{P}_1)=20$, that is to say when there is no stochasticity in the training (black curve).
    We furthermore observe a slight reduction of the inference bias as $\redu{\operatorname{Card}}(\mathcal{P}_1)$ augments.
    This illustrates how the theoretical bias of the PAVI-E scheme identified in \cref{sec:no_bias} does not translate in significantly worse empirical results ---albeit in the pathological case of $\redu{\operatorname{Card}}(\mathcal{P}_1)=1$.
    Interestingly, the training speed is approximately constant across all values of $\redu{\operatorname{Card}}(\mathcal{P}_1)$.
    This feature is essential in the PAVI-E scheme: we can train over a reduced version of our model ---with a light memory footprint--- and apply the obtained architecture to the full model.
    Note that this feature wouldn't necessarily be present when training on CPU: computing summary statistics over larger sets would make the training \textit{slower} when $\redu{\operatorname{Card}}(\mathcal{P}_1)$ augments.
    }}
    \label{fig:card_redu}
\end{figure}

In \cref{fig:card_redu} we show the impact of the augmentation of $\redu{\operatorname{Card}}(\mathcal{P}_1)$ on the efficiency of the variational posterior's optimization.

In practice, we noticed that the most efficient choice in the case of the PAVI-F scheme was to maximize the cardinalities of the reduced model given the memory constraints of the GPU.
Indeed, training over larger cardinalities does make each optimization step slightly slower but also makes the ELBO gradient estimates less noisy and allows to train more encoding vectors $\tE_{i,n}$ at a given optimization step.

In the case of the PAVI-E scheme, the training speed is constant with respect to $\redu{\operatorname{Card}}(\mathcal{P}_1)$.
This is  due to the computing of the encodings being vectorized across plates in our deep-set encoder ---see \cref{sec:PAVI-E_details}~\citep{ST}.
We observe a slight if barely noticeable reduction of the inference bias when augmenting $\redu{\operatorname{Card}}(\mathcal{P}_1)$.
Similar to the PAVI-F scheme, $\redu{\operatorname{Card}}(\mathcal{P}_1)$ should be maximized with respect to the GPU memory constraint.
The intuition behind this choice is that the generalization of the learning of summary statistics across sets of data points is easier the closer the reduced set size is to the full set size.
This constant training speed allows for a controlled memory footprint of the stochastic training: contrary to the PAVI-F scheme, $\redu{\operatorname{Card}}(\mathcal{P}_1)$ could be maintained at a value manageable by the GPU without reducing the training speed.
This example also displays a pathological choice for $\redu{\operatorname{Card}}(\mathcal{P}_1)$: the encoder fails to learn to compute the correct summary statistics over sets of size 1.
However trivial, this example underlines that $\redu{\operatorname{Card}}(\mathcal{P}_1)$ should be chosen at a value logical with respect to the posterior's sufficient statistics.
When generalizing to moments of a higher order ---such as the variance--- there is therefore a theoretical lower bound to consider when fixating the value of $\redu{\operatorname{Card}}(\mathcal{P}_1)$.
Our intuition is that the more complex to estimate the statistic is, the larger $\redu{\operatorname{Card}}(\mathcal{P}_1)$ should be.

Note that in practice non-stochastic VI is intractable for large-scale models, because of its memory requirement.
In large-scale experiments such as the one presented in \cref{fig:full}, stochastic training is necessary.
Our claim is to be faster than non-plate-amortized stochastic VI in those large-scale contexts ---but not necessarily to be faster than non-stochastic VI in the small-scale regime of this experiment.
For PAVI-F we nonetheless obtain similar training speed compared to non-stochastic VI with $\redu{\operatorname{Card}}(\mathcal{P}_1)$ as small as $8$.
For PAVI-E, the stochastic training is as fast as the non-stochastic one.

\subsection{Additional comparative experiments}
\label{sec:sup_comparisons}
\subsubsection{Gaussian Mixture model}
\label{sec:exp_gm}

In this experiment, we test out various baselines over a challenging model: a Gaussian mixture with random effects.
\begin{equation}
\label{eq:GM}
\begin{aligned}
    D,\quad \operatorname{Card}(\mathcal{P}_1), \quad \operatorname{Card}(\mathcal{P}_0) &= 2,\quad 20,\quad 10 \\
    \substack{\forall n_1=1..\operatorname{Card}(\mathcal{P}_1) \\ \forall n_0=1..\operatorname{Card}(\mathcal{P}_0)} \quad  X_{n_1,n_0} |\theta_{1,n_1}, \pi_{n_1} &\sim \operatorname{Mixture}(\left[ \mathcal{N}(\theta_{1,n_1}^1, \sigma_x^2), \hdots , \mathcal{N}(\theta_{1,n_1}^L, \sigma_x^2) \right], \pi_{n_1}) \\
    \forall n_1=1..\operatorname{Card}(\mathcal{P}_1) \quad \pi_{n_1} &\sim \operatorname{Dirichlet}(1 \times \vec 1_L) \\
    \substack{\forall n_1=1..\operatorname{Card}(\mathcal{P}_1) \\ \forall l=1..L} \quad \theta_{1,n_1}^l | \theta_{2,0}^l &\sim \mathcal{N}(\theta_{2,0}^l, \sigma_1^2) \\
    \forall l=1..L \quad \theta_{2,0}^l &\sim \mathcal{N}(\vec 0_D, \sigma_2^2) \enspace ,
\end{aligned}
\end{equation}
where $\operatorname{Mixture}([\mathcal{D}_1, \hdots , \mathcal{D}_L], \pi)$ denotes a mixture between the distributions $[\mathcal{D}_1, \hdots , \mathcal{D}_L]$ with mixture weights $\pi$.
\rebutal{Results are visible in \cref{tab:GM}, where PAVI displays the best asymptotic ELBO, as well as the shortest optimization time.}
On that note, we underline that stochastic training over a mixture distribution is challenging, as ---due to the sub-sampling of points--- only a fraction of the mixture components could be expressed at a given step, requiring the architecture to dynamically cluster the data points across time.

\begin{table}
    \centering
    \begin{tabular}{rll}
        Architecture & ELBO ($10^1$) & Optimization time (s) \\
        \\ \hline \\
        CF & - 16.2 ($\pm$ 2.9) & 3,100 \\
        ADAVI & - 20.0 ($\pm$ 2.3) & 3,000 \\
        \rebutal{UIVI} & - 26.3 ($\pm$ 1.2) & 500 \\
        PAVI-E (ours) & - 15.2 ($\pm$2.8) & 470 \\
        PAVI-F (ours) & \textbf{- 11.1 ($\pm$1.1)} & \textbf{300}
    \end{tabular}
    \caption{{\color{new}\textbf{Analytical performance over a Gaussian mixture HBM} PAVI shows superior performance. PAVI converges in a fraction of CF's optimization time ---the second-best-performing architecture.}}
    \label{tab:GM}
\end{table}

\subsubsection{Hierarchical Variance model}
\label{sec:HV}
In this experiment, we test out our architectures over a non-canonical model, in which parent RVs play the role of variance for the distribution of their children.
Our goal is in particular to evaluate a potential empirical bias for the PAVI-E scheme ---as studied in \cref{sec:no_bias}.
We will refer to the following HBM as the Hierarchical Variance model:
\begin{equation}
    \label{eq:HV}
    \begin{aligned}
    D,\quad \operatorname{Card}(\mathcal{P}_1), \quad \operatorname{Card}(\mathcal{P}_0) &= 2,\quad 15,\quad 15 \\
    \substack{\forall n_1=1..\operatorname{Card}(\mathcal{P}_1) \\ \forall n_0=1..\operatorname{Card}(\mathcal{P}_0)} \quad  \log X_{(n_1,n_0)} |\theta_{1,n_1} &\sim \mathcal{N}(0, \theta_{1,n_1}) \\
    \forall n_1=1..\operatorname{Card}(\mathcal{P}_1) \quad \log \theta_{1,(n_1)} | \theta_{2,0} &\sim \mathcal{N}(0, \theta_{0,0}) \\
    \log \theta_{2,0} &\sim \mathcal{N}(\vec 0_D, 1) \enspace
    \end{aligned}
\end{equation}
In this model, the encoder ---used in the PAVI-E scheme--- has to collect non-trivial summary statistics: empirical variances across variable subsets of the observed data.
This is the context in which we would expect to observe the most bias due to the approximation in \cref{eq:approx_PAVI-E}.
The performance of PAVI-E remains nonetheless competitive.
This illustrates how PAVI-E's theoretical bias ---introduced in the stochastic training--- does not result in significantly worse inference compared to state-of-the-art architectures.
\rebutal{PAVI also displays the best ELBO, in conjunction with UIVI, but in a 5 times shorter optimization time.}

\begin{table}
    \centering
    \begin{tabular}{rll}
        Architecture & ELBO ($10^2$) & Optimization time (s) \\
        \\ \hline \\
        CF & - 11.2 ($\pm$ 2.9) & 500 \\
        \rebutal{UIVI} & -6.7 ($\pm$ 0.8) & 100 \\
        PAVI-E (ours) & -6.7 ($\pm$1.0)$^1$ & 90 \\
        PAVI-F (ours) & \textbf{- 6.7 ($\pm$0.8)} & \textbf{20} \\
    \end{tabular}
    \caption{{\color{new}\textbf{Analytical performance over a Hierarchical Variance HBM} In this setting with hard-to-estimate summary statistics, PAVI-E doesn't show any empirical bias.
    $^1$: PAVI-E suffers from numerical instability in this example, with runs degenerating into NaN results.}}
    \label{tab:HV}
\end{table}

\subsubsection{Small dimension version of our Neuroimaging model}
\label{sec:exp_toy_HCPL}

In this experiment, we validate the performance of our architecture on synthetic data generated using a small dimension version of the model presented in \cref{eq:HCPL}.
To allow the use of the comparative baselines, we fixate $S, \redu{S} = 5, 3$, $T, \redu{T} = 6, 3$, $N, \redu{N} = 12, 3$, $D=4$, $L=7$.
\rebutal{Results are visible in \cref{tab:HCPL}: PAVI can deal with HBMs unavailable to the ADAVI architecture \citep{ADAVI}, and so so with a performance superior to the CF baseline \citep{CF}.
Despite intensive efforts, we did not manage to obtain a numerically stable UIVI optimization.
This is likely due to the complexity of this inference problem: a large dimensionality (2k parameters) combined with complex dependencies between RVs in the posterior.
In this context, automatic structured VI baselines such as CF or PAVI exploit the parametric form of the prior $p$ to help with inference.
PAVI's plate amortization likely facilitates the inference, resulting in a reduced \textit{optimization error}~\citep{bottou_tradeoffs_2007}.}

\begin{table}
    \centering
    \begin{tabular}{rll}
        Architecture & ELBO ($10^4$) & Optimization time (s) \\
        \\ \hline \\
        CF & - 139.7 ($\pm$ 21) & 3,200 \\
        ADAVI & --- & --- \\
        \rebutal{UIVI} & --- & --- \\
        PAVI-F (ours) & \textbf{- 8.2} ($\pm$ \textbf{0.1}) & \textbf{1,600}
    \end{tabular}
    \caption{{\color{new}\textbf{Analytical performance over our Neuroimaging HBM}
    This HBM is not pyramidal and consequently cannot be processed by the ADAVI architecture.
    \rebutal{Due to the large parameter space (approximately 2k parameters) and the complex density involved, despite intensive efforts we did not manage to obtain a numerically stable UIVI optimization.}
    PAVI yields a higher ELBO than CF, in half of CF's optimization time.}}
    \label{tab:HCPL}
\end{table}
}

\subsection{Experimental details - analytical examples}
\label{sec:details_arch}

All experiments were performed in Python, using the \textit{Tensorflow Probability} library \citep{dillon_tensorflow_2017}.
Through this section, we refer to \textit{Masked Autoregressive Flows} \citep{papamakarios_masked_2018} as \textit{MAF}.
All experiments are performed using the Adam optimizer \citep{adam_kingma2014}.
At training, the ELBO was estimated using a Monte Carlo procedure with $8$ samples.
All architectures were evaluated over a fixed set of $20$ samples $\tX$, with 5 seeds per sample.
Non-sample-amortized architectures were trained and evaluated on each of those points.
Sample amortized architectures were trained over a dataset of $20,000$ samples separate from the 20 validation samples, then evaluated over the 20 validation samples.

\tmlrrebutal{\textbf{Termination condition}
Our termination condition amounts to plateau detection across the statistics of multiple runs. In details, we ran the inference with a large, conservative number of epochs for all methods and collected the evolution of the ELBO for each run. We then took as runtime the time to reach an ELBO close to the one at convergence for the method. We averaged over 100 repetitions to avoid stochastic noise due to Monte Carlo estimation of the ELBO. The order of magnitude difference between the results is clearly above the potential noise effects due to our procedure.}

\subsubsection{Plate amortization and convergence speed (\ref{sec:exp_convergence_speed})}

All 3 architectures (baseline, PAVI-F, PAVI-E) used:
\begin{itemize}
    \item for the flows $\mathcal{F}_i$, a MAF with $\left[ 32, 32 \right]$ hidden units;
    \item as encoding size, $128$
\end{itemize}
For the encoder $f$ in the PAVI-E scheme, we used a multi-head architecture with 4 heads of 32 units each, 2 ISAB blocks with 64 inducing points.

\subsubsection{Impact of encoding size (\ref{sec:exp_encoding_size})}

All architectures used:
\begin{itemize}
    \item for the flows $\mathcal{F}_i$, a MAF with $\left[ 32, 32 \right]$ hidden units, after an affine block with triangular scaling matrix.
    \item as encoding size, a value varying from $2$ to $16$
\end{itemize}

\subsubsection{Scaling with plate cardinalities (\ref{sec:exp_scaling})}

\textbf{ADAVI} \citep{ADAVI} we used:
\begin{itemize}
\item for the flows $\mathcal{F}_i$, a MAF with $\left[ 32, 32 \right]$ hidden units, after an affine block with triangular scaling matrix.
\item for the encoder, an encoding size of $8$ with a multi-head architecture with 2 heads of 4 units each, 2 ISAB blocks with 32 inducing points.
\end{itemize}

\textbf{Cascading Flows} \citep{CF} we used:
\begin{itemize}
    \item a mean-field distribution over the auxiliary variables $r$
    \item as auxiliary size, a fixed value of 8
    \item as flows, \textit{Highway Flows} as designed by the Cascading Flows authors
\end{itemize}

\textbf{PAVI-F} we used:
\begin{itemize}
    \item for the flows $\mathcal{F}_i$, a MAF with $\left[ 32, 32 \right]$ hidden units, after an affine block with triangular scaling matrix.
    \item an encoding size of 8
\end{itemize}

\textbf{PAVI-E} we used:
\begin{itemize}
    \item for the flows $\mathcal{F}_i$, a MAF with $\left[ 32, 32 \right]$ hidden units, after an affine block with triangular scaling matrix.
    \item for the encoder, an encoding size of $16$ with a multi-head architecture with 2 heads of 8 units each, 2 ISAB blocks with 64 inducing points.
\end{itemize}

\rebutal{\textbf{UIVI} we used as $\operatorname{Card}(\mathcal{P}_1) = 2 \rightarrow 20 \rightarrow 200$:
\begin{itemize}
    \item as base distribution, a standard Gaussian with dimensionality $6 \rightarrow 42 \rightarrow 402$
    \item as transform $h$, an affine transform with diagonal scale
    \item as embedding distribution, a standard Gaussian with dimensionality $3 \rightarrow 6 \rightarrow 9$
    \item as transform weights regressor, an MLP with hidden units $[32, 32] \rightarrow [64, 64] \rightarrow [128, 128]$
    \item to sample uncorrelated samples $\epsilon$, an HMC run with 5 burn-in steps and 5 samples
    \item an Adam optimizer with exponential learning rate decay, starting at $1e-2$, $\times 0.9$ every 300 steps
\end{itemize}}

\subsubsection{Gaussian mixture (\ref{sec:exp_gm})}

\textbf{ADAVI} \citep{ADAVI} we used:
\begin{itemize}
\item for the flows $\mathcal{F}_i$, a MAF with $\left[ 32 \right]$ hidden units, after an affine block with diagonal scaling matrix.
\item for the encoder, an encoding size of $16$ with a multi-head architecture with 2 heads of 8 units each, 2 ISAB blocks with 8 inducing points.
\end{itemize}

\textbf{Cascading Flows} \citep{CF} we used:
\begin{itemize}
    \item a mean-field distribution over the auxiliary variables $r$
    \item as auxiliary size, a fixed value of 16
    \item as flows, \textit{Highway Flows} as designed by the Cascading Flows authors
\end{itemize}

\textbf{PAVI-F} we used:
\begin{itemize}
    \item for the flows $\mathcal{F}_i$, a MAF with $\left[ 32 \right]$ hidden units, after an affine block with diagonal scaling matrix.
    \item an encoding size of 16
    \item $\redu{\operatorname{Card}}(\mathcal{P}_1)=5$
\end{itemize}

\textbf{PAVI-E} we used:
\begin{itemize}
    \item for the flows $\mathcal{F}_i$, a MAF with $\left[ 128, 128 \right]$ hidden units, after an affine block with triangular scaling matrix.
    \item for the encoder, an encoding size of $128$ with a multi-head architecture with 4 heads of 32 units each, 2 ISAB blocks with 128 inducing points.
    \item $\redu{\operatorname{Card}}(\mathcal{P}_1)=5$
\end{itemize}

\rebutal{\textbf{UIVI} we used:
\begin{itemize}
    \item as base distribution, a standard Gaussian with dimensionality $82$
    \item as transform $h$, an affine transform with diagonal scale
    \item as embedding distribution, a standard Gaussian with dimensionality $6$
    \item as transform weights regressor, a MLP with hidden units $[64, 64]$
    \item to sample uncorrelated samples $\epsilon$, an HMC run with 5 burn-in steps and 5 samples
    \item an Adam optimizer with exponential learning rate decay, starting at $1e-2$, $\times 0.9$ every 300 steps
\end{itemize}}

\subsubsection{Hierarchical Variances (\ref{sec:HV})}

\textbf{Cascading Flows} \citep{CF} we used:
\begin{itemize}
    \item a mean-field distribution over the auxiliary variables $r$
    \item as auxiliary size, a fixed value of 32
    \item as flows, \textit{Highway Flows} as designed by the Cascading Flows authors
\end{itemize}

\textbf{PAVI-F} we used:
\begin{itemize}
    \item for the flows $\mathcal{F}_i$, a MAF with $\left[ 32, 32 \right]$ hidden units, after an affine block with diagonal scaling matrix.
    \item an encoding size of 32
    \item $\redu{\operatorname{Card}}(\mathcal{P}_1)=3$
    \item $\redu{\operatorname{Card}}(\mathcal{P}_0)=3$
\end{itemize}

\textbf{PAVI-E} we used:
\begin{itemize}
    \item for the flows $\mathcal{F}_i$, a MAF with $\left[ 128, 128 \right]$ hidden units, after an affine block with diagonal scaling matrix.
    \item for the encoder, an encoding size of $128$ with a multi-head architecture with 4 heads of 32 units each, 2 ISAB blocks with 128 inducing points.
    \item $\redu{\operatorname{Card}}(\mathcal{P}_1)=3$
    \item $\redu{\operatorname{Card}}(\mathcal{P}_0)=3$
\end{itemize}

\rebutal{\textbf{UIVI} we used:
\begin{itemize}
    \item as base distribution, a standard Gaussian with dimensionality $32$
    \item as transform $h$, an affine transform with diagonal scale
    \item as embedding distribution, a standard Gaussian with dimensionality $3$
    \item as transform weights regressor, an MLP with hidden units $[64, 64]$
    \item to sample uncorrelated samples $\epsilon$, an HMC run with 5 burn-in steps and 5 samples
    \item an Adam optimizer with exponential learning rate decay, starting at $1e-2$, $\times 0.9$ every 300 steps
\end{itemize}}

\subsubsection{Small Neuroimaging example (\ref{sec:exp_toy_HCPL})}

\textbf{Cascading Flows} \citep{CF} we used:
\begin{itemize}
    \item a mean-field distribution over the auxiliary variables $r$
    \item as auxiliary size, a fixed value of 32
    \item as flows, \textit{Highway Flows} as designed by the Cascading Flows authors
\end{itemize}

\textbf{PAVI-F} we used:
\begin{itemize}
    \item for the flows $\mathcal{F}_i$, a MAF with $\left[ 32, 32 \right]$ hidden units, after an affine block with diagonal scaling matrix.
    \item an combination of encoding sizes of 32 and 8
\end{itemize}

\rebutal{\textbf{UIVI} we used:
\begin{itemize}
    \item as base distribution, a standard Gaussian with dimensionality $1,802$
    \item as transform $h$, an affine transform with diagonal scale
    \item as embedding distribution, a standard Gaussian with dimensionality $16$
    \item as transform weights regressor, an MLP with hidden units $[128, 256, 512, 1024]$
    \item to sample uncorrelated samples $\epsilon$, an HMC run with 5 burn-in steps and 5 samples
    \item an Adam optimizer with exponential learning rate decay, starting at $1e-3$, $\times 0.9$ every 300 steps
\end{itemize}
Optimization systematically degenerated into NaN results after around $200$ optimization steps.}

\subsection{Details about our Neuroimaging experiment (\ref{sec:exp_fMRI})}
\label{sec:details_neuro}
\subsubsection{Data description}
\label{sec:fMRI_data}
In this experiment, we use data from the \textit{Human Connectome Project (HCP)} dataset \citep{HCP}.
We randomly select a cohort of $S=1,000$ subjects from this dataset, each subject is associated with $T=2$ resting-state fMRI sessions \citep{rfMRI}.
We minimally pre-process the signal using the \texttt{nilearn} python library \citep{nilearn}:
\begin{enumerate}
    \item removing high variance confounds
    \item detrending the data
    \item band-filtering the data (0.01 to 0.1 Hz), with a repetition time of 0.74 seconds
    \item spatially smoothing the data with a 4mm Full-Width at Half Maximum
\end{enumerate}

For every subject, we extract the surface Blood Oxygenation Level Dependent (BOLD) signal of $N=59,412$ vertices across the whole cortex.
We compare this signal with the extracted signal of $D=64$ DiFuMo components: a dictionary of brain spatial maps allowing for an effective fMRI dimensionality reduction \citep{difumo}.
Specifically, we compute the one-to-one Pearson's correlation coefficient of every vertex with every DiFuMo component.
The resulting connectome, with $S$ subjects, $T$ sessions, $N$ vertices and a connectivity signal with $D$ dimensions, is of shape $(S \times T \times N \times D)$.
We project this data ---correlation coefficients lying in $\left] -1; 1 \right[$--- in an unbounded space using an inverse sigmoid function.

\subsubsection{Model description}
We use a model inspired by the work of \citet{kong_spatial_2018}.
We hypothesize that every vertex in the cortex belongs to either one of $L=7$ functional networks.
This number is inspired by the work of \citet{parc_1} and \citet{kong_spatial_2018}, in which authors identify $7$ general networks on resting-state fMRI data.

Each network corresponds to a pattern of connectivity with the brain, represented as the correlation of the BOLD signal with the signal from the $D=64$ DiFuMo components.
We define $L=7$ functional networks at the population level.
Each network $l$ corresponds to the connectivity fingerprint $\mu_{l}$.

At the population level, for a vertex $n$ we denote as $\text{logits}_{n} \in \mathrm{R}^L$ the logits of its probability to belong to each network $l$.
Each subject $s$ is associated with the logits $\text{logits}_{s, n}$, which are considered as a perturbation of the population logits.
This creates a regularization across subjects: the same vertex $n$ is encouraged to have the same label across all subjects.
The variable $\gamma_l$ controls the inter-subject spatial variability across all subjects, for the network $l$.

For every subject $s$, session $t$, and vertex $n$, we denote as $X_{s, t, n}$ the observed connectivity.
This connectivity is modeled via a mixture model: $X_{s, t, n}$ is a perturbation of either one of the $\mu_l$'s connectivity fingerprints.
$\text{label}_{s, n}$ denotes the label in the mixture (that depends on the subject-specific logits).
$\kappa_l$ controls the variability between $X_{s, t, n}$ and the mixture component $\mu_{\text{label}_{s,n}}$.

The resulting model can be described as:
\begin{equation}
\label{eq:HCPL}
    \begin{aligned}
    S, T, N, D, L &= 1000, 2, 59412, 64, 7 \\
    \substack{\forall l=1..L}: \quad \mu_{l} &\sim \mathcal{N}(0 \times \vec 1_D, 6) \\
    \substack{\forall l=1..L}: \quad \log \kappa_{l} &\sim \mathcal{N}(0 \times \vec 1_L, 1) \\
    \substack{\forall n=1..N}: \quad \operatorname{logits}_{n} &\sim \mathcal{N}(0 \times \vec 1_L, 6) \\
    \substack{\forall l=1..L}: \quad \log \gamma_{l} &\sim \mathcal{N}(0 \times \vec 1_L, 1) \\
    \substack{\forall s=1..S \\ \forall n=1..N}: \quad \operatorname{logits}_{s, n} | \operatorname{logits}_{n}, [\gamma_l]_{l=1..L} &\sim \mathcal{N}(\operatorname{logits}_{n}, [\gamma_1 ... \gamma_L]) \\
     \substack{\forall s=1..S \\ \forall n=1..N}: \quad \operatorname{labels}_{s, n} | \operatorname{logits}_{s, n} &\sim \operatorname{Categorical}(\operatorname{logits}_{s, n}) \\
    \substack{\forall s=1..S \\ \forall t=1..T \\ \forall n=1..N}: \quad X_{s,t,n} | [\mu_{l}]_{l=1..L}, [\kappa_{l}]_{l=1..L}, \operatorname{label}_{s,n} &\sim \mathcal{N}(\mu_{\operatorname{label}_{s,n}}, \kappa_{\operatorname{label}_{s,n}})
    \end{aligned}
\end{equation}
The model contains 4 plates: the \textit{network} plate of full cardinality $L$ (that we did not exploit in our implementation), the \textit{subject} plate of full cardinality $S$, the \textit{session} plate of full cardinality $T$ and the \textit{vertex} plate of full cardinality $N$.

Our goal is to recover the posterior distribution of the networks $\mu$ and the labels $\operatorname{label}$ ---represented as the parcellation in \cref{fig:full}--- given the observed connectome described in \cref{sec:fMRI_data}.

\subsubsection{PAVI implementation}

We used in this experiment the PAVI-F scheme, using:
\begin{itemize}
    \item for the RVs $\mu_l$:
    \begin{itemize}
        \item for the flows $\mathcal{F}_i$, a MAF with $\left[ 64, 128, 64 \right]$ hidden units, following an affine block with diagonal scale
        \item for the encoding size: $16$
    \end{itemize}
    \item for the RVs $\kappa_l, \gamma_l$:
    \begin{itemize}
        \item a MAP estimator using a regressor with $\left[ 64, 128, 64 \right]$ hidden units
        \item for the encoding size: $8$
    \end{itemize}
    \item for the RVs $\operatorname{logits}_{n}, \operatorname{logits}_{s, n}$:
    \begin{itemize}
        \item for the flows $\mathcal{F}_i$, an affine block with diagonal scale regressed using a $\left[ 64, 128, 64 \right]$ MLP
        \item for the encoding size: $8$
    \end{itemize}
    \item for the reduced model, we used $\redu{S}=128$, $\redu{T}=1$ and $\redu{N}=128$.
\end{itemize}

To allow for the optimization over the discrete $\operatorname{label}_{s,n}$ RV, we used the Gumbell-Softmax trick, using a fixed temperature of $1.0$ \citep{gumbell_softmax, concrete}.

\subsubsection{Cognitive scores prediction}

For the cognitive scores prediction in \cref{fig:full}, we reproduced the standard methodology from \citet{kong_spatial_2018}.

We perform a 20-fold cross-validation across 1,000 subjects.
We start from the $\operatorname{logits}_{s, n}$ associated with each subject.
We use PCA to project the features from the 19 training folds to their $100$ first components (with 33\% explained variance).
We then train a linear regression to predict each of the 13 cognitive scores from the training-fold PCA features.
We compute the test performance on the test fold ---using the training set PCA and linear regression--- averaged across the 13 cognitive measures.
This process is reproduced 100 times.
The reported scores are the triple average across folds, measures, and repetition, while the standard deviation is computed across the 100 repetitions only.

\section{Supplemental discussion}
\subsection{Plate amortization as a generalization of sample amortization}
\label{sec:sup_sa}
In \cref{sec:plate_amortization}, we introduced plate amortization as applying the generic concept of amortization to the granularity of plates.
There is actually an even stronger connection between sample amortization and plate amortization.
 
An HBM $p$ models the distribution of a given observed RV $X$ ---jointly with the parameters $\Theta$.
Different samples $\tX_0, \tX_1, ...$ of the model $p$ are i.i.d. draws from the distribution $p(X)$.
$p$ can thus be considered as the model for "one sample".
Instead of $p$, consider a "macro" model for the whole \textit{population} of samples one could draw from $p$.
The observed RV of that macro model would be the infinite collection of samples drawn from the same distribution $p(X)$.
In that light, the i.i.d. sampling of different $X$ values from $p$ could be interpreted as a plate of the macro model.
Thus, we could consider sample amortization as an instance of plate amortization for the "sample plate".
Or equivalently: plate amortization can be seen as the generalization of amortization beyond the particular case of sample amortization.

\subsection{Alternate formalism for SVI --- PAVI-E(sa) scheme}
In this work, we propose a different formalism for SVI, based around the concept of full HBM $\mathcal{M}^{\text{full}}$ versus reduced HBM $\mathcal{M}^{\text{redu}}$ sharing the same template $\mathcal{T}$.
This formalism is helpful to set up GPU-accelerated stochastic VI \citep{dillon_tensorflow_2017}, as it entitles a fixed computation graph -with the cardinality of the reduced model $\mathcal{M}^{\text{redu}}$- in which encodings are "plugged in" -either sliced from larger encoding arrays or as the output of an encoder applied to a data slice, see \cref{sec:variational_family}\&\ref{sec:sharing_learning}.
Particularly, our formalism doesn't entitle a control flow over models and distributions, which can be hurtful in the context of \textit{compiled} computation graphs such as in \textit{Tensorflow} \citep{tensorflow2015-whitepaper}.

The reduced model formalism is also meaningful in the PAVI-E(sa), where we train and amortized variational posterior over $\mathcal{M}^{\text{redu}}$ and obtain "for free" a variational posterior for the full model $\mathcal{M}^{\text{full}}$ ---see \cref{sec:sharing_learning}.
In this context, our scheme is no longer a different take on hierarchical, batched SVI: the cardinality of the full model is truly independent of the cardinality of the training and is only simulated as a scaling factor in the stochastic training ---see \cref{sec:stochastic_training}.
We have the intuition that fruitful research directions could stem from this concept.

\subsection{Benefiting from structure in inference}
Our contributions can be abstracted through the concept of plate amortization -see \cref{sec:plate_amortization}.
\rebutal{Plate amortization is particularly useful in the context of heavily parameterized density approximators such as normalizing flows, but is not tied to it: plate-amortized Mean Field \citep{blei_variational_2017}, ASVI \citep{ASVI}, or implicit \citep{SIVI, UIVI} schemes are also possible to use.}
Plate amortization can be viewed as the amortization of common density approximators across different sub-structures of a problem.
This general concept could have applications in other highly-structured problem classes such as graphs or sequences \citep{wu_comprehensive_2020, rnn_review}.

Central to our design, PAVI adjoins an encoding structure to an inference problem.
The encodings $\tE_{i,n}$ embed all the "individualized" information in the problem, while the shared conditional density estimators $\mathcal{F}_i$ translate those encodings into posterior distributions.
We believe there is potential in externalizing part of the inference problem into an embedding.
For instance, embeddings could be learned separately from density estimators, opening up the possibility for transfer learning~\citep{weiss_survey_2016}.
Or the encodings could integrate some known symmetry of the problem, such as convolution-based encodings in the case of Random Fields~\citep{book_bishop}.

\rebutal{\subsection{Connection with meta-learning}
In \cref{sec:plate_amortization}, we introduced plate amortization: sharing the parameterization and learning across a model's plates.
This section discusses the connection between plate amortization and meta-learning\citep{ABML,meta_learning_amo_VI, meta_learning_hierch_struct}.

Supervised learning can be seen as the mapping from a given context set $\mathcal{C}=\{ (x,y) \}$ to a predictive function $f$~\citep{book_bishop} such that $f(x) = y$.
Meta-learning ---or "learning to learn"--- instead recovers this mapping $\mathcal{C} \mapsto f$ in the general case.
Once the meta-training is completed, a predictive function $f$ conditioned by an unseen context $\mathcal{C}$ can be obtained in a single forward pass ---without any training done on $\mathcal{C}$.
As an instance of meta-learning, the Neural Process Family encodes the context $\mathcal{C}$ via a deep set encoder~\citep{NPF_1, NPF_2, deep_sets}.
The encoded context, along with the data point $x$ is then used to condition an estimator for the density $q(y | x, \mathcal{C})$.

This framework is similar to the PAVI-E scheme, where a combination of a deep set encoder and a normalizing-flow-based density estimator output the posterior probability of a ground RV $\theta_{i,n}$.
This encoder-estimator pair is repeatedly used across a plate.
This is analogous to meta-learning to solve the inference problem across the different elements of a plate ---such as the different subjects of a population study.
In PAVI-E, the encoding $\tE_{i,n}$ at a lower hierarchy plays a role similar to the context $\mathcal{C}$ in meta-leaning.
A few differences, however, exist between the two frameworks:
\begin{itemize}
    \item meta-learning is typically concatenated to the 1-plate regime, whereas PAVI is designed for the multi-hierarchy scenario;
    \item meta-learning typically considers a set of i.i.d. tasks, whereas in PAVI the inference of different ground RVs $\theta_{i,n}$ are conditionally dependent through the hierarchical model $p$;
    \item meta-learning is trained using the forward KL loss, that is to say in the sample-amortized regime, maximizing the probability $q(y)$ of samples from the underlying generative process. In contrast, PAVI ---though possible to train using the forward KL loss--- is trained via the reverse KL, which requires to evaluate the density of the generative process explicitly.
\end{itemize}
We suspect there would be interesting applications for the PAVI architecture in hierarchical meta-learning scenarios.}

\subsection{Conditional dependencies modeled in the variational family}
\label{sec:dep_modelled}
Consider an inference problem with three latent parameters $\theta_1$, $\theta_2$, $\theta_3$, and a model $p(\Theta, X)$ that factorizes as $p(\theta_1, \theta_2, \theta_3) = p(\theta_1)p(\theta_2)p(\theta_3 | \theta_1,\theta_2)p(X|\theta_3)$.
To approximate the posterior $p(\Theta | X)$, different conditional dependencies can be modeled in the chosen variational family $\mathcal{Q}$.
The mean-field (MF) approximation fully factorizes the variational distribution as $q(\theta_1, \theta_2, \theta_3) = q(\theta_1) q(\theta_2)q(\theta_3)$.
MF was originally introduced in VI to facilitate computation, allowing dedicated optimization schemes \citep{blei_variational_2017}.
Nonetheless, this approximation assumes independence between RVs in the posterior and ultimately limits the expressivity of the variational family.
To remove this approximation, modeling complex conditional dependencies in the variational family is an open research subject \citep{CF, webb_faithful_2018}.
In the PAVI design, we inherit our statistical dependency structure from the prior distribution $p$, as detailed in \cref{eq:q_full}.
This amounts to factorizing $q(\theta_1, \theta_2, \theta_3) = q(\theta_1)q(\theta_2)q(\theta_3|\theta_1,\theta_2)$.
This choice of dependencies follows the line of research from \citet{hoffman_structured_2014} and \citet{ASVI}.
As pointed out by \citet{CF}, when modeling only those \textit{forward} dependencies, the modeling of colliders can be an issue.
More generally, one would like to model arbitrary dependencies in the variational posterior.
As an example, the factorization $q(\theta_1, \theta_2, \theta_3) = q(\theta_3)q(\theta_2|\theta_3)q(\theta_1|\theta_2,\theta_3)$ would more faithfully represent the conditional dependencies that can arise in the posterior.
Modeling arbitrary dependencies reduces the inference gap but can become computationally intractable as the number of RVs increases.
In PAVI, we strike a particular balance between expressivity and computational tractability, as further motivated in this section.

In the case of a single plate ---the 2-level case--- \citet{agrawal_amortized_2021} demonstrate that modeling only the \textit{forward} dependencies does not result in reduced expressivity compared to the modeling of the full dependencies.
Yet this result does not hold in the n-level case, as \citet{webb_faithful_2018} shows that faithful inversion features conditional dependencies in the posterior between ground RVs of the same template.
We can dub those dependencies as \textit{horizontal} dependencies across RVs in the same plate.
Like Structured SVI \citep{hoffman_structured_2014} or Automatic SSVI \citep{ASVI}, the PAVI design thus results in reduced expressivity when stacking multiple plates in the model $p$.
This issue can be partially alleviated with the usage of a \textit{backward} encoding scheme ---going in the reverse direction compared to the prior's dependencies--- as in the PAVI-E design (see \cref{sec:sharing_learning}) or in Cascading Flows \citep{CF}.

In practice, though limiting our expressivity, \textit{horizontal} dependencies are difficult to inject back into our architecture.
Critically, in the PAVI design, the use of a common density estimator across the ground RVs of the same template (see \cref{sec:archi}) and the stochastic training over batches of those RVs (see \cref{sec:stochastic_training}) prevent the direct modeling \textit{horizontal} dependencies.
Put differently, the fact that we consider the inference over different ground RVs as conditionally independent inference problems is central to our design and adverse to the modeling of \textit{horizontal} dependencies.

Injecting \textit{horizontal} dependencies back into our variational family is therefore a non-trivial research direction, that is not at the core of this paper.
This opens up promising research directions: how could arbitrary conditional dependencies be modeled in the variational posterior in the context of stochastic training?

\subsection{Towards user-friendly Variational Inference}
By re-purposing the concept of amortization at the plate level, we aim to propose clear computation versus precision trade-offs in VI.
Hyper-parameters such as the encoding size ---as illustrated in \cref{fig:encoding_size} (right)--- allow to clearly trade inference quality in exchange for a reduced memory footprint.
On the contrary, in classical VI, changing $\mathcal{Q}$'s parametric form ---for instance, switching from Gaussian to Student distributions--- can have a strong and complex impact on the number of weights and inference quality \citep{blei_variational_2017}.
By allowing the usage of normalizing flows in very large cardinality regimes, our contribution aims at disentangling approximation power and computational feasibility.
In particular, having access to expressive density approximators for the posterior can help experimenters diversify the proposed HBMs, removing the need for properties such as conjugacy to obtain meaningful inference \citep{Gelman_book}.
Combining clear hyper-parameters and scalable yet universal density approximators, we tend towards a user-friendly methodology in the context of large population studies VI.

\tmlrrebutal{
In its current implementation, PAVI requires to set up a rather large number of hyper-parameters.
This includes choosing a normalizing flow architecture, the size of the encodings at different plate levels, and the reduced model cardinalities.
Those hyper-parameters are in addition to generic ones such as the number of samples from the variational family used to estimate the ELBO, or the choice of an optimizer.
However, our experiments suggest that those hyper-parameters could default to conservative values.
As shown in \cref{sec:exp_encoding_size}, the encoding size should be maximized with respect to the available memory, as should the reduced cardinalities, as detailed in \cref{sec:exp_card_redu}.
As for the normalizing flow architecture, since the latter is shared across plates, the user can always default to the most expressive architecture available without incurring an exploding number of weights.
As an example, in our experiments, we stuck to the MAF architecture~\citep{papamakarios_fast_2018}.
Since all hyper-parameters can be set conservatively, a user-friendly automatic VI API can be designed, where the user only provides the system with the generative model $p$, and some observed value $\tX$.
The variational family $q$ can then automatically be derived from $p$, and directly optimized over $\tX$.
This is the design principle we implemented in the codebase adjoined to this publication.
}

\end{document}